\documentclass[sigplan,authorversion,nonacm]{acmart}
\usepackage{graphicx}
\usepackage{amsfonts}
\usepackage{tabularx}
\usepackage{bm}

\usepackage{dsfont}
\usepackage{multirow}
\usepackage{balance}

\usepackage{algorithm}
\usepackage{algorithmic}
\usepackage{balance}

\usepackage{algorithm}
\usepackage{algorithmic}
\AtBeginDocument{%
  \providecommand\BibTeX{{%
    \normalfont B\kern-0.5em{\scshape i\kern-0.25em b}\kern-0.8em\TeX}}}

\begin{document}

\title{Quantitative Evaluations on Saliency Methods: An Experimental Study}

%
\author{Xiao-Hui Li}
\orcid{0000-0003-4561-9096}
\affiliation{%
  \institution{Huawei Technologies}
  \city{Shenzhen}
  \state{China}
  \postcode{518129}
}
\email{lixiaohui33@huawei.com}

\author{Yuhan Shi}
\affiliation{%
  \institution{Huawei Technologies}
  \city{Shenzhen}
  \country{China}
}
\email{shiyuhan1@huawei.com}

\author{Haoyang Li}
\affiliation{%
  \institution{Huawei Technologies}
  \city{Hong Kong SAR}
  \country{China}
}
\email{li.haoyang@huawei.com}

\author{Wei Bai}
\affiliation{%
  \institution{Huawei Technologies}
  \city{Hong Kong SAR}
  \country{China}
}
\email{baiwei16@huawei.com}

\author{Yuanwei Song}
\affiliation{%
  \institution{Huawei Technologies}
  \city{Hong Kong SAR}
  \country{China}
}
\email{song.yuanwei@huawei.com}

\author{Caleb Chen Cao}
\orcid{0000-0001-5109-3700}
\affiliation{%
  \institution{Huawei Technologies}
  \city{Hong Kong SAR}
  \country{China}
}
\email{caleb.cao@huawei.com}

\author{Lei Chen}
\affiliation{%
  \institution{Hong Kong University of Science and Technology}
  \city{Hong Kong SAR}
  \country{China}
}
\email{leichen@cse.ust.hk}

\renewcommand{\shortauthors}{Li and Shi, et al.}


\begin{abstract}
It has been long debated that eXplainable AI (XAI) is an important topic, but it lacks rigorous definition and fair metrics. In this paper, we briefly summarize the status quo of the metrics, along with an exhaustive experimental study based on them, including  faithfulness, localization, false-positives, sensitivity check, and stability. With the experimental results, we conclude that among all the methods we compare, no single explanation method dominates others in all metrics. Nonetheless, Gradient-weighted
Class Activation Mapping (Grad-CAM) and Randomly Input Sampling for Explanation (RISE) perform fairly well in most of the metrics. Utilizing a set of filtered metrics, we further present a case study to diagnose the classification bases for models. While providing a comprehensive experimental study of metrics,  we also examine measuring factors that are missed in current metrics and hope this valuable work could serve as a guide for future research. 

\end{abstract}



\maketitle

\section{Introduction}\label{sec:Introduction}
{A}{rtificial} Intelligence (AI) technology has gained tremendous development in the past decade, especially in computer vision (CV) community \cite{voulodimos2018deep}. Since the birth of AlexNet \cite{krizhevsky2012imagenet}, deep learning (DL) based methods have outperformed traditional methods and can even beat human beings in many open challenges (e.g. ImageNet image-classification challenge \cite{ILSVRC15}, COCO object-detection challenge \cite{lin2014microsoft}). However, the high non-linearity and complexity make these models black-boxes,  i.e., it is difficult for humans to understand their internal working mechanisms and decision-making processes. Such intransparency has raised concerns not only among the related industries but also among governments and organizations. 

As a result, XAI has emerged as an important and popular research direction among academia in the past few years \cite{yang2019evaluating, li2020survey, adadi2018peeking}. An exponentially increasing number of scientific works have been conducted to push forward this branch of research \cite{selvaraju2017grad, lundberg_unified_2017, ordookhanians2019demonstration, boer2020personal, sellam2019deepbase}.
In CV scenario, a very natural way to provide explanations is through visualization. Among visualization, one of the most popular branches is through generating \textit{saliency map} that shows the importance of input features towards the final output on the pixel/super-pixel level, e.g. Grad-CAM \cite{selvaraju2017grad} and SHapley Additive exPlanations (SHAP) \cite{lundberg_unified_2017}, Local Interpretable Model-agnostic Explanations (LIME) \cite{ribeiro2016should} .  Compared to other visualization methods such as feature visualization \cite{nguyen2016synthesizing}, saliency methods are usually simpler and easier to deploy. Besides, the explanation of saliency map is within the input space where the value of the saliency map at a specific position directly indicates the importance of that pixel and is thus usually more intuitive.

As tools explaining the decision-making process, XAI methods can be applied in human-in-the-loop machine learning systems \cite{qian2019systemer, parameswaran2019enabling, krishnan2017palm} or model debugging systems \cite{ma2018mode, vartak2018mistique,sellam2019deepbase}. 
However, to deploy XAI methods in practical systems, a methodological challenge appears: \textit{how to evaluate these methods such that the users can choose the most suitable one for their needs}. 

From the perspective of experiments conduction for accessing XAI methods, the corresponding evaluation metrics can be grouped into human-grounded metrics and functionally-grounded ones \cite{doshi2017towards}. Human-grounded metrics are usually designed through human-involving experiments, e.g. designing questionnaires for the explanations and invite a group of tester to answer them. Collecting their answers and a quantified evaluation result can be obtained \cite{hoffman2018metrics}.  However, it is difficult to scale up such experiments for big data scenario as human participants can be costive and hard to standardize. In this sense, functionally-grounded metrics that do not involve human testers might be a better choice for large scale data. Along this path, different metrics have been proposed to quantify different properties of XAI methods.

A few works quantified the \textbf{\textit{faithfulness}} of explanations. \cite{samek2016evaluating, petsiuk2018rise} proposed area over/under perturbation curve as metrics of faithfulness when removing/inserting important features the explainer indicates. Some alternative works quantified the \textbf{\textit{localization}} ability of saliency methods. Simonyan \textit{et al.} \cite{zhang2018top} proposed Pointing Game (PG) which counts the ratio of data samples in which the pixel with the highest relevance score lies in the bounding box.  Starting from constructing controllable "ground-truth" for explanations, Yang and Kim \cite{yang2019bim} proposed three metrics to evaluate \textbf{\textit{false-positives}} of explanations based this dataset:  Model Contrast Score (MCS), Input dependence Rate (IDR) and Input Independence Rate (IIR). Some other works studied if XAI methods have certain desired sensitivities (we term it \textit{\textbf{sensitivity check}}). \cite{adebayo2018sanity} proposed to check whether the XAI methods are sensitive to model weights and \cite{rebuffi2020there} proposed to check the class sensitivity of the XAI methods. In contrast to desired sensitivities, there are works propose to check the \textbf{\textit{stability}} of XAI methods towards undesired variations (e.g., insignificant noise on the input),  \cite{melis2018towards, dombrowski2019explanations, yeh2019fidelity} proposed different but similar metrics by perturbing the input slightly and measure the corresponding difference on the explanation. 

Although these metrics have been developed, an overall quantitative comparison with all these metrics among the widely used XAI methods, e.g. Grad-CAM, has not yet been conducted.  In this work, we aim to fill in this gap by performing extensive experiments to compare these different saliency methods with the proposed metrics.  As we start from large scale datasets, in this paper, we only consider computational metrics as mentioned above.  Hopefully, our results can serve as a guide to choose a suitable method for a given use-case. Our main contributions are as follows:

\begin{itemize}
\item A thorough quantitative evaluation based on different metrics is conducted to compare some of the widely deployed XAI methods that generate saliency maps as explanations. With the experimental results, we conclude the pros and cons of these methods.

\item We provide strategic augmentation of the benchmark dataset \cite{yang2019bim} to adapt to more objects and more scenes. 
\item We modify the PG metric for localization and the MCS metric for false-positives to enhance their testing usability.
\item We present a case study of the XAI methods and metrics in actual model analysis, which helps diagnose the spurious correlations in the dataset.
\end{itemize}

\textbf{Outline.} This rest of this paper is organized as the following: we first review some of the popular and widely discussed XAI methods that generate saliency maps. After that, we introduce the definition and calculation of the proposed evaluation metrics, summarizing the advantages and disadvantages of them as evaluation metrics. Based on these metrics, we conduct an exhaustive comparative experiment on the methods reviewed in Section. \ref{sec:method} with different benchmark datasets and black-box models. The experimental results and findings are reported in Section. \ref{sec:experiment}. Following the experimental results, we present a simple utilization of the metrics and XAI methods in Section \ref{sec:utilization}.  Finally, we conclude and highlight the future work in Section. \ref{sec:conclusion}. 
 
\section{Saliency Methods}\label{sec:method}
Saliency methods are popular explainers, especially in image classification scenario. These explainers quantify the "importance" of individual pixels w.r.t the final out and generate a visualized saliency map in
pixel/input space. From the mechanism of computing such "importance", these methods can be grouped into two categories: 1. Backpropagation methods: the relevance is calculated through backpropagation; 2. Perturbation-based methods: the importance of a certain pixel is quantified as the output drop when perturbing this pixel. 
With the efforts of scientists in this community, plenty of innovative methods have been proposed to generate high-quality saliency maps. In this section, we review some of the most widely used methods according to the aforementioned categories.

In the following descriptions, an input image is represented as  $I \in \mathbb{R}^{ch\times h\times w} $, where $ch,h,w$ are the number of channels, height, and width of the input image. A black-box classifier can be described as a function $f: \mathbb{R}^{ch\times h\times w}\rightarrow \mathbb{R}^{C}$, where $C$ is the number of classes in the image classification problem. An explanation method that generates the saliency map for the input image and black-box model is represented as a functional $E:(\mathbb{R}^{ch\times h\times w}, f)\rightarrow \mathbb{R}^{h\times w}$, where the output saliency map has the same spatial dimension as the input image. 

\subsection{Backpropagation Methods}
The basic idea of backpropagation methods is to compute class-relevance of individual pixels by propagating the quantity of interest (e.g. gradients, relevance, excitation) back to the input image space. 
The claim is that the higher the value of the propagated signal at a particular pixel, the more important that pixel is to the final output.

\subsubsection{Gradient}

Proposed by Simonyan \textit{et al.,} Gradient directly computes the gradients of the output class w.r.t the input pixels of a given image as the explanation \cite{simonyan2013deep}. Specifically, following the definition in \cite{simonyan2013deep}, the saliency map for class $ c $  can be calculated as follows :
\begin{equation}
 E_{Gradient}(I, f)_c = \frac{\partial f_c(I')}{\partial I'} \bigg|_{I'=I}.
\end{equation}
This formulation can also be understood as: $ f_c(I') $ can be approximated with a linear function in the neighborhood of $ I$ by the first-order Taylor expansion: $ f_c(I') \approx  E(I, f)_c^T\cdot (I'-I) + b $. The weights of this approximated linear function are the generated saliency map.


%

\subsubsection{Guided Backpropagation}

Despite the simplicity of Gradient method, the generated saliency maps are usually noisy and difficult to understand. To overcome this issue, Springenberg \textit{et al.} proposed Guided Backpropagation (GBP) with an additional guidance signal from the higher layers \cite{springenberg2014striving}. It prevents backward flow of negative gradients, which decreases the activation of the higher layer unit. Denoting $ \{A_0, A_1, …A_k\} $ as the feature maps during the forward pass at each layer, and $ \{R_0, R_1, … R_k\}  $ as the signal obtained in the backward pass,
GBP zeros out negative gradients during the backpropagation of $ R $ as follows:
\begin{equation}\label{eq:GBP}
\textrm{GBP}: R^{i} = (A^{i} > 0) \cdot (R^{i+1} > 0) \cdot R^{i+1}
\end{equation}
where $ (A^{i} > 0) $ keeps only the positive activations, and $ (R^{k+1} > 0) $ keeps the positive gradients only. Starting from the last layer and iterating Equation. (\ref{eq:GBP}) until $i=0$, the saliency map can then be obtained as $E_{GBP}(I, f) = R^0$.

\subsubsection{Integrated Gradients}

Gradient method suffers from another issue called "gradient saturation" that gradient vanishes with certain value of input and cannot represents the sensitivity of this input.
Integrated gradients (IntGrad) \cite{sundararajan1703axiomatic} addresses this issue by summing over scaled versions of inputs. Introducing a baseline input $I'$, IntGrads are defined as the path integral of the gradient along the straight line path from the baseline $ I' $ to the input $ I $:
\begin{equation}
E_{IntGrad}(I, f) = (I - I') \times \int_{\alpha = 0}^{1} \frac{\partial f(I' + \alpha \times (I - I'))}{\partial I} d\alpha
\end{equation}
For images, the baseline could be an image with constant value or an image respecting an empirical distribution.

\subsubsection{ Grad-CAM}

\begin{figure}[htp]
	\centering
	\includegraphics[width = 8.3cm]{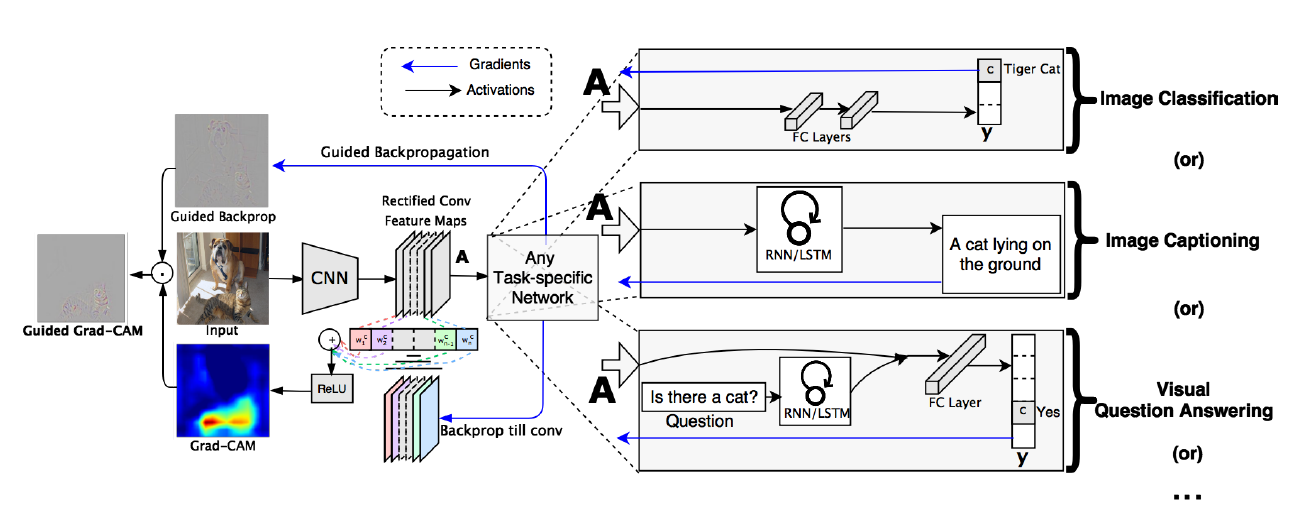}
	\caption{Grad-CAM Overview : Given an image and a target class, the image is fed into the network and the activation maps for the layers of interest are obtained. A one-hot signal with desired class set to 1 is then backpropagated to the rectified convolutional feature maps of interest, which is combined to compute the coarse Grad-CAM saliency map. \cite{selvaraju2017grad} }
	\label{fig:gradcam}
\end{figure}

Introduced by Selvaraju \textit{et al.,} Grad-CAM uses the class-specific gradient flowing into the last convolutional layer of a CNN model (rather than flowing back to input as in Gradient method) to produce a coarse localization map of the important regions in the image \cite{selvaraju2017grad}. It is a generalization of Class Activation Mapping (CAM) \cite{zhou2016learning} where CAM requires global average pooling layer on the fully CNN models while Grad-CAM can be applied to CNN models with fully connected layers. As shown in Figure. \ref{fig:gradcam}, in order to get the class-discriminative localization map, the first step is to compute the gradient of the score for class $ c $, $ y^c $ (before softmax) with respect to feature maps $ A^k $ of a convolutional layer. Then the neuron importance weights are calculated:

\begin{equation}
a^{c}_{k} = \frac{1}{Z}\sum_{i}\sum_{j}\frac{\partial y^{c}}{\partial A^{k}_{ij}}
\end{equation}
The weight $ a^{c}_{k} $ represents the importance of feature map $ A^k $ for a specific class $ c $. Then, performing a weighted combination of forward activation maps followed by a ReLU operation, saliency maps are generated through:
\begin{equation}
E_{Grad-CAM}(I, f)_c = \text{ReLU}(\sum_{k} a^{c}_{k}A^{k})
\end{equation}
 The application of ReLU here is to focus on features with positive influence on the class of interest. In most of the use-cases, the feature maps $A^k$ are set to be the activation maps of the last convolutional layer. 

%

\subsubsection{Contrastive Excitation Backpropagation}
Like Grad-CAM, Excitation Backpropagation \cite{zhang2018top} also generates interpretable attention maps at intermediate convolutional layers. Different from the former methods that back propagate the gradients, Excitatation Backpropagation defines a top-down propagation rule through excitatory connections between activation neurons as follows:
\begin{equation}\label{eq:ebp}
P(a_j | a_i)  = \begin{cases}
Z_i \hat{a}_j w_{ji} \quad  &\text{if} \quad w_{ji}\geq 0\\
0 \quad \quad \quad &\text{Otherwise}
\end{cases}
\end{equation}
where $ \hat{a}_{j} $ is the activation function coming from lower layer, 
$ w_{ij} $ is the weight from neuron $ i $ to neuron $ j $. $Z_i = 1/\sum_{j:w_{ji}\geq 0}\hat{a}_j w_{ji}$ is a normalization factor. Saliency map can be computed from any intermediate convolutional layers by recursively propagating the top-down signal layer by layer through Equation. (\ref{eq:ebp}), then taking the sum across channels.
Contrastive Excitation Backpropagation (CEBP) inherits the idea of Excitation Backpropagation and adopts the idea of contrasting the excitation of one class with ones of all the others. This further enhances the class discriminative maps. To do this, given an output unit $ o_i \in O $, a dual unit  $ \overline{o_{i}} \in O $ is virtually constructed, whose input weights are the negation of those of $ o_i $. For example, if an output unit
corresponds to an "elephant" classifier, then its dual unit will correspond to a
non-"elephant" classifier.
By subtracting the saliency map for $ \overline{o_i} $ from the one for $ o_i $ will cancel out common winner neurons and enhanced the discriminative neurons. Finally, the result is contrastive saliency map, which better highlights the regions making the image unique to the target class.

\subsection{Perturbation-based Methods}
The aforementioned backpropagation methods require access to the internals of the model to be explained, such as the gradients of the output w.r.t the input, intermediate feature maps, or the network's weights. On the contrary, perturbation-based methods are truly "black-box" methods. The saliency map is generated by perturbing the input and observing its effects on the output. Without actual access to the structures or internal states of the models, perturbation-based methods are therefore model-agnostic.

\subsubsection{Occlusion}\label{subsubsec:Occlusion}

Occlusion \cite{zeiler2014visualizing} is a visualization technique that answering if the black-box model is truly identifying the location of the object in the image or just using the surrounding
context. Specifically, it systematically replaces different contiguous rectangular regions of the input image with a baseline input value, and monitors how the feature maps and classifier outputs changes. The  saliency map for a specific class $c$ can be calculated by Equation. (\ref{eq:occlusion}):
\begin{equation} \label{eq:occlusion}
E_{Occlusion}(I, f)_c = \sum_{k}\frac{f_c(I) - f_c({I}_{k}^{'})}{m\times m} \mathbb{I}_k
\end{equation}
where $ {I}_{k}^{'} $ is an perturbed image masked by an $m\times m$ square and $\mathbb{I}_k$ is an indicator function that for pixels within the square region $\mathbb{I}_k=1$, otherwise $\mathbb{I}_k=0$. For features located in multiple regions, the corresponding output differences are averaged to compute the attribution for that feature. 

\subsubsection{RISE}\label{subsubsec:RISE}

\begin{figure}[htp]
	\centering
	\includegraphics[width = 8.3cm]{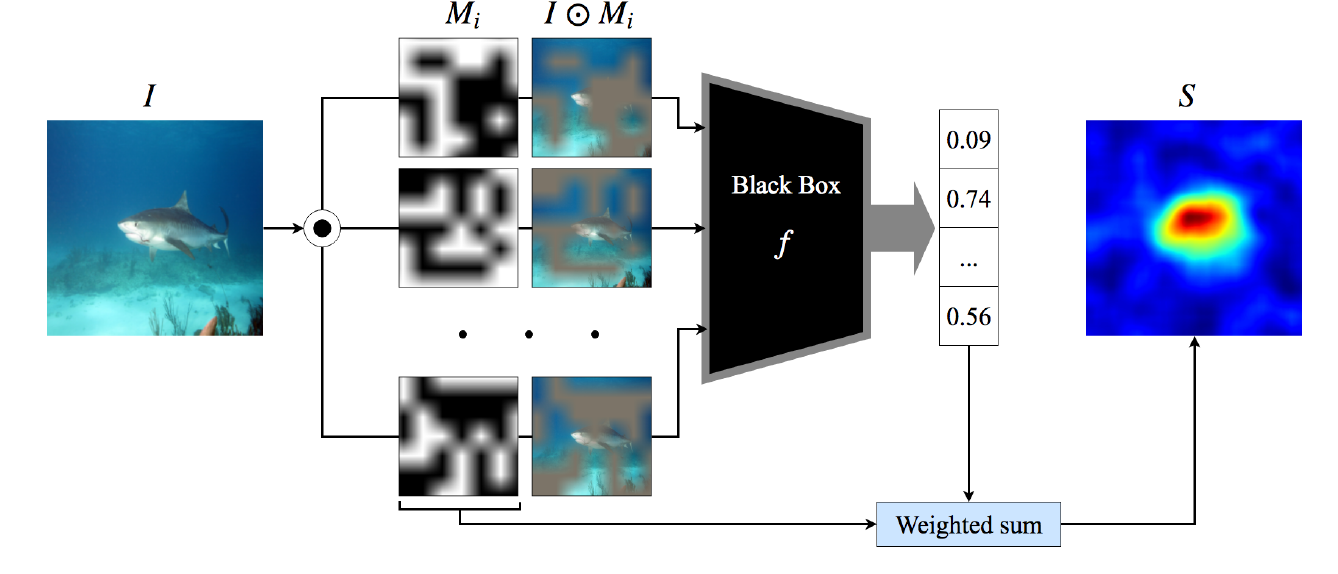}
	\caption{RISE Overview: The input image $ I $ is element-wise multiplied with random masks $ M_{i} $ and the masked images are fed to the base model. The saliency map is a linear combination of the masks where the weights are the score of the target class corresponding to the respective masked inputs \cite{zhang2018top} }
	\label{fig:rise.PNG}
\end{figure}

Randomized Input Sampling for Explanation (RISE) generates saliency maps by sampling on multiple random binary masks. As shown in Figure. \ref{fig:rise.PNG}, the original image is randomly masked, and then fed into the black-box model and gets a prediction. The final saliency map is the weighted sum of these random masks, with the weights being the corresponding output on the node of interest:
\begin{equation}
E_{RISE}(I, f)_c = \sum_{i}f_c(I\odot M_i)  M_i 
\end{equation}
where $c$ is the class of interest, $M_i$ is the random mask and $\odot$ is element-wise product in spatial dimensions.  The idea behind this is that if a mask preserves important parts of the image, it gets a higher score on the output, and consequently has a higher weight and a more dominant contribution on the final saliency map.

%
%
\section{Evaluation Metrics}\label{sec:metric}
As mentioned in the Section. \ref{sec:Introduction}, there are different works proposed to assess the XAI methods from different dimensions:  faithfulness, localization, false-positives, sensitivity check, and stability. In this section, we review the representative metrics designed for these dimensions and present the detailed mathematical formulations which serve as the guideline for the exact implementations.     

\subsection{Faithfulness}\label{subsec:Faithfulness}
Saliency methods generates relevance scores w.r.t model predictions assigned to the features (pixels or super-pixels for images). Described in \cite{melis2018towards}, the faithfulness of an explanation refers to whether the relevance scores reflect the true importance. A typical approach for quantifying the property of XAI is through strategical modification on the input according to the explainer indication and monitor the model behaviors. There are several different metrics proposed based on this approach. In this paper, we implement the metric iAUC (termed as $insertion$ in the original paper) proposed in \cite{petsiuk2018rise} and conduct the experiments with it.

$\bullet$\textit{\textbf{iAUC}} Petsiuk \textit{et al.} \cite{petsiuk2018rise} proposed area under the insertion curve (iAUC), which accumulated the probability increase while gradually inserting the features from the original input to a reference input ( e.g. a constant-value image or a blurred image). The features from the original input were inserted in the order of high to low relevance score indicated by the explanation method. 
 Given a reference input $I_r$, the origin input $I$, and the indexes of features $\{q_0, q_1, q_2..., q_L\}
$ that sort the relevance scores from high to low, the calculation of iAUC for a single input can be described as:
\begin{equation}
\begin{split}
I_{p_0} & = I_r \\ 
 I_{p_{k+1}}[q_k] & = I[q_k],\  \mathrm{for}\  \{k=0, ...., L \} \\
\mathrm{iAUC} & = \frac{1}{L+1} \sum_{i=1}^{L+1} (f_c(I_{p_i}) - f_c (I_r))
\end{split}
\end{equation}
where $f_c(\cdot)$ is the model output on class $c$. Faithful explanations are expected to obtain violent increase at the beginning, and thus the greater AUC.

Faithfulness metrics have the advantages that no extra annotations are needed and the idea of using model behavior changes as a signal of quantifying the explanations is intuitive enough. However, they have a critical issue in the actual calculation: the problem of data shifts. The modifications to the inputs can drag the data out of the learned distribution of the black-box model. Consequently, an enormous change could happen even if low-relevance pixels are removed. This causes unreasonable results when computing the faithfulness of saliency maps, thus further efforts are needed to tackle this issue.

\subsection{Localization}\label{subsubsec:human_annotation}
For XAI methods that generate relevance scores for features, one of the criteria for the explanations to be good is that it correctly recognize the discriminative features for the model. However, it is hard to find the "ground-truth"  discriminative features of a black-box model. In object recognition tasks, a natural assumption is that a well-trained black-box model would make predictions based on the features from the object itself. With this assumption, the evaluation relaxes to whether the explainer correctly localizes the object to be recognized. 


$\bullet$\textit{\textbf{PG}} Zhang \textit{et al.} \cite{zhang2018top} proposed PG to evaluate the spatial localization capability w.r.t different labels. Taking the images from COCO \cite{lin2014microsoft} and PASCAL VOC 2007 \cite{pascal-voc-2007}, the evaluation is simply based on the maximum pixel from the saliency map. If the pixel with the highest score locates inside the ground truth area, it counts as a hit otherwise a miss. Iterating all the images, one can get the number of hits and misses, PG can be simply computed through:  
\begin{equation}
\mathrm{PG} = \frac{\#Hits}{\#Hits+\#Misses}.
\end{equation}
It should be emphasized that the localization of a saliency map refers to the locating capability of a saliency map. Since saliency methods are originally designed to find the features most relevant to model predictions, the saliency map might highlight a few discriminative super-pixels which are considered critical for the decision rather than the whole object. Therefore, a good saliency map is not required to cover the entire object. 

\subsection{False-positives}
As a critical difficulty for evaluating the saliency map is that no ground truth saliency map is available for comparison, an interesting line of works proposed to construct controllable datasets and produce pseudo-ground truth for evaluations \cite{kim2017interpretability, yang2019bim}.  Specifically, Yang and Kim \cite{yang2019bim} constructed Benchmarking Attribution Methods (BAM) dataset by combining objects from Microsoft COCO \cite{lin2014microsoft} and scene images from Miniplaces \cite{zhou2017places}, and proposed three metrics to evaluate the false-positives of XAI methods: Model Contrast Score (MCS), Input dependence Rate (IDR) and Input Independence Rate (IIR). Since IIR can be considered as an extension of stability metric which we shall discuss in the next subsection, we only discuss MCS and IDR here. Taking the definition from \cite{yang2019bim}, the average contribution of a specific concept $co$ on a single image is calculated as:

\begin{equation}
\label{eq:CF_contribution}
	S_{co}(f,I_0) = \frac{1}{\sum M_{co}} \sum E(f,I_0) \odot M_{co}
\end{equation}
where $M_{co}\in\{0, 1\}^{h\times w}$ represents the binary mask where pixels of concept $co$ take value 1, and 0 otherwise. The black-box model and input image are denoted as $f$ and $I_0$. The contribution $S_{co}$ takes average on the Hadamard product of saliency map and binary mask.

Global concept contribution $G_{co}(\cdot)$ is further defined as averaged $S_{co}$ over all the correctly predicted images $\mathbb{I}_{corr}$:

\begin{equation}\label{eq:global_CF_contribution}
	G_{co}(f, \mathbb{I}) = \frac{1}{|\mathbb{I}_{corr}|}\sum_{I_0 \in \mathbb{I}_{corr}}{S_{co}(f, I_0)}
\end{equation}

$\bullet$ \textit{\textbf{MCS}}
Two models are trained on the same dataset of BAM but with two sets of labels $L_o$ and $L_s$. Object labels $L_o$ are used to train the object classifier $f_o$, while scene labels $L_s$ are used to train scene classifier $f_s$. The object commonly appearing in every scene image is considered as a common feature (CF). For the object classifier $f_o$, CFs are important features and thus should be assigned higher scores. On the contrary, scene classifier $f_s$ is expected to rely on scenic features rather than the CFs. Since CFs are more important to object classifier $f_o$ than scene classifier $f_s$, global contribution of CF, $G_{cf}(f_o, \mathbb{I})$ are expected larger than $G_{cf}(f_s, \mathbb{I})$. MCS is defined as the difference between $G_{cf}(f_o, \mathbb{I})$ and $G_{cf}(f_s, \mathbb{I})$.

\begin{equation}\label{eq:MCS}
	\mathrm{MCS} = G_{cf}(f_o, \mathbb{I}^{cf}) - G_{cf}(f_s, \mathbb{I}^{cf})
\end{equation}
MCS measures how differently the CFs are attributed to different classifiers. Object classifiers are supposed to make decisions based on CF, and hence, better explanation should obtain higher MCS.

$\bullet$ \textit{\textbf{IDR}}
Considering the scene classifier $f_s$, if there are two images with the same scenery but with and without CF, the region covered by CF in the image ought to have a smaller contribution to the model prediction than the same area in its counterpart image without CF. 
IDR calculates the proportion of saliency maps that assign high scores to CF as follows:
\begin{equation}
	\label{eq:FPR}
		\mathrm{IDR} = \frac{1}{|\mathbb{I}^{cf}|} \sum_{0}^{k}{\mathds{1} ((S_{cf}, I^{cf}_k) > (S_{cf}, I^{\neg cf}_k))}
\end{equation}
where $I^{cf}_k \in \mathbb{I}^{cf}$ and $I^{\neg cf}_k \in \mathbb{I}^{\neg cf}$ are the images with and without the CF. For two images with same scenery, average contribution of pixels covered by CF in the image with CF is compared to the contribution of the same region of its counterpart image without CF. If the former is greater the latter, it counts as a hit. IDR is obtained as the hit rate over the entire dataset, indicating the proportion of images where CFs are assigned high importance scores. Better explanation methods should have smaller IDR.

The constructed datasets provide controllable pseudo-ground-truths for explanations and thus can serve as benchmark datasets for quantifying the faithfulness of explanation methods. However, the generalization of such dataset is questionable: the behaviors of explanation methods on different data and  different black-box models can be different, thus the conclusion of the evaluations of explanation methods on the synthetic datasets may not be able to directly generalize to other datasets.

\subsection{Sensitivity Check}\label{subsec:sensitivity_check}

Besides the attempts to directly design metrics to quantify specific properties of  explanation methods, there are works focusing on checking whether the explanation methods have desired sensitivities. 


$\bullet$ \textit{\textbf{Class Sensitivity}} is one of the desired sensitivities. If an explanation method is faithful, it should be giving different explanations for different decisions. Take image classification as an example, most of the XAI methods provide saliency maps that highlight the discriminative part as explanations. Different classes usually have different discriminative regions, thus a good explanation method should have clear class sensitivity. Guillaumin \textit{et al.} \cite{guillaumin2012large} took classes with the highest and the lowest confidence and computed the similarity between their saliency maps as a metric for class sensitivity (without confusion, we denote the metric as CS in the following text). Given an input $I$, a classifer $f$ and an explanation method $E$, CS metric can be then defined as:
\begin{equation}
\begin{split}
c_{max}, c_{min} &= \arg\max f(I), \arg\min f(I) \\
\mathrm{CS} &= \mathrm{sim} (E(I, f)_{c_{max}}, E(I, f)_{c_{min}})
\end{split}
\end{equation}
where $\mathrm{sim}(\cdot, \cdot)$ is a similarity measurement function, e.g. Pearson correlation. 

For a good explanation method, the saliency maps between the classes with the highest score and the lowest score should be very different since these two classes usually do not have similar discrimination strategies. Thus a good explanation method should have CS close to zero or below zero.    

We note here, these checks can only be applied as a necessity check, e.g. for a good explanation method, CS must be low, but it does not necessarily work the other way around.

\subsection{Stability}\label{subsec:Robustness}
In Section. \ref{subsec:sensitivity_check}, we review the works that check whether the XAI methods have the desired sensitivities. Starting from a contrary aspect, some alternative works attempt to evaluate the stability (insensitivity) of XAI methods towards insignificant variations. A good explanation is expected to be stable when an input is perturbed slightly but still has a similar model prediction and confidence distribution. If perceptibly identical images with similar model outputs have completely different saliency maps, it could cause problems in applications. For example, in autonomous driving scenario, passengers may receive completely different explanations, even if there is no obvious change within a few seconds. They will get confused and lose trust in the car system. 
 

$\bullet$\textit{\textbf{Max-sensitivity}}  
 Yeh \textit{et al.} put forward a simple quantity termed max-sensitivity to measure the stability of an explanation method through the following definition \cite{yeh2019fidelity}:
\begin{equation}
\mathrm{SENS_{max}}(E,f, I, r) = \max_{\parallel \delta \parallel <r} \parallel E(f, I+\delta) -E(f, I) \parallel 
\end{equation}
where $r$ is a predefined parameter indicating the range of perturbation. The attraction of this quantity is that it can be robustly estimated via Monte-Carlo sampling. Another advantage of this quantity is its simplicity in the sense of implementation. In the experiments, we applied this metric definition to quantify the stability of the selected saliency methods.

\section{Experiments}\label{sec:experiment}
To give an overall comparison of the popular methods mentioned in Section. \ref{sec:method}, for each dimension discussed in Section. \ref{sec:metric}, we choose one or two metrics to conduct a set of comparative experiments. All of the experiments are performed across the validation/test set of the two benchmark datasets: Microsoft COCO 2017 \cite{lin2014microsoft} and PASCAL VOC 2012 \cite{pascal-voc-2007}, except for false-positives, which we generated an extended BAM (eBAM) dataset to evaluate. For each dataset, two types of pretrained models are explained: ResNet50 \cite{he2016deep} and VGG16 \cite{simonyan2014very}. 
For all the metrics, the overall results are obtained as the average through all the samples in dataset. 

\subsection{Faithfulness Results} \label{subsec:experiment_faithfulness}

\begin{figure}[h!]
\centering
\includegraphics[width=1\linewidth]{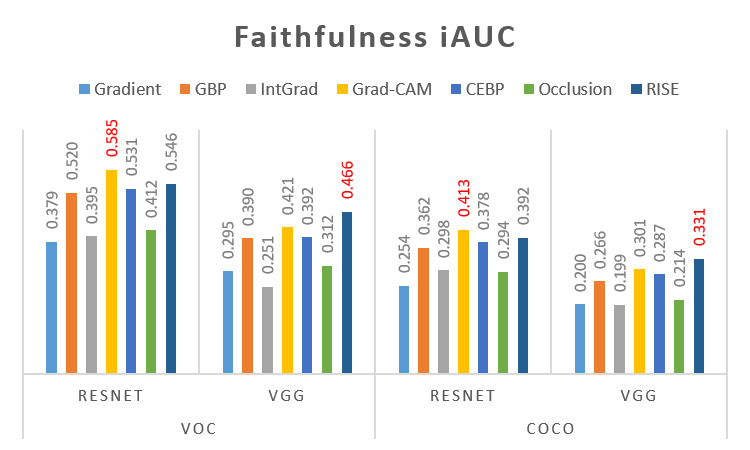}
\caption{Faithfulness results with iAUC metric across VOC and COCO datasets, with the best methods highlighted. Different saliency methods are represented by different colors. The number above every bar is the overall iAUC.  }\label{fig:faithfulness_results}
\end{figure}

Following the definition of iAUC and the setting described in \cite{petsiuk2018rise}, we conduct the faithfulness experiments with the blurred images set as the perturbed reference images. As argued in \cite{petsiuk2018rise}, compared to setting the reference images as constant-value images, the blurred images could ease the issue of sharp shapes introduction during the insertion.  The overall results are shown in Figure. \ref{fig:faithfulness_results}.

 From Figure. \ref{fig:faithfulness_results}, one can see that: (1) RISE and Grad-CAM perform the best in datasets VOC and COCO.  Specifically, RISE performs the best for
VGG16 in both datasets and Grad-CAM gets the best results for ResNet50. The reason might be that ResNet50 has deeper structure than VGG16 and thus the feature maps from the last convolutional
layer of ResNet50 might be more semantic and class discriminative;  (2) The ranking of faithfulness for the selected methods is: Grad-CAM $\approx$ RISE > CEBP > GBP > Occlusion > IntGrad $\approx$ Gradient.

\begin{figure*}[ht!]
\centering
\includegraphics[width=0.8\linewidth]{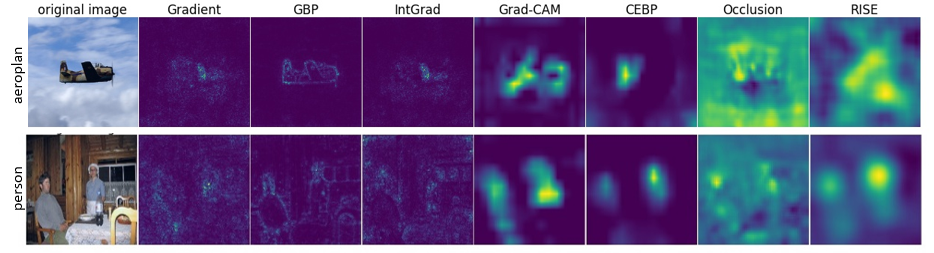}
\caption{Typical saliency maps of the listed methods. Data samples are from PASCAL VOC and the pretrained model is VGG16.}\label{fig:faithness_saliency}
\end{figure*}

To inspect the results, we visualize a few examples in Figure. \ref{fig:faithness_saliency} of the generated saliency maps. The visualization shows that: Grad-CAM indeed captures the objects to be recognized. RISE has similar salient regions as Grad-CAM, but it has diffusive noise in the background. The salient region of CEBP can be seen as a subregion of that of Grad-CAM. Gradient, IntGrad and Occlusion have
noisy saliency maps while GBP highlights the edges of the objects
clearly.

The reason that Grad-CAM performs better might be that it calculates the attribution of the last convolutional layer, which better capture high-level semantic information. And the reason that RISE performs better that Occlusion might be that the perturbation
on RISE are more flexible than Occlusion (does not require to be patch-based perturbation), thus the sampling reservoir can be larger and the sampling results are less noisy and more faithful. IntGrad and Gradient perform the worst because they include negative gradients and the saliency maps are noisier.
 

\begin{figure}[ht]
\centering
\includegraphics[width=0.8\linewidth]{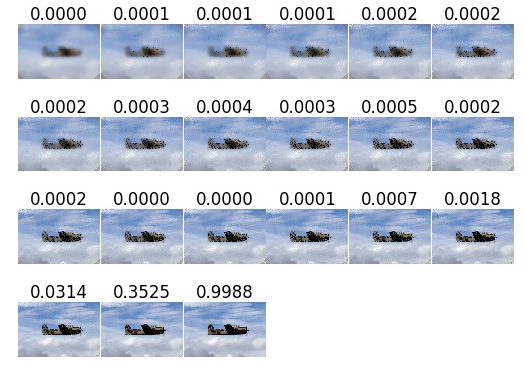}
\caption{An example that shows the intermediate process of iAUC. The number above each subfigure is the corresponding output probability of the node of interest ("aeroplane" in this example). From upper left to lower right, more and more pixels are inserted in the order of indicated importance of saliency maps.}
\vspace{-4ex}\label{fig:faithness_gradient}
\end{figure} 

Although we chose iAUC as the metric for our faithfulness experiments, the out of distribution issue is still a severe problem. As shown in Figure. \ref{fig:faithness_gradient}, the output probability of the third to last subfigure is closed to zero while the last subfigure contains a few more pixels and the output probability is nearly one. 

\subsection{Localization Results}
\begin{figure}[h!]
\centering
\includegraphics[width=\linewidth]{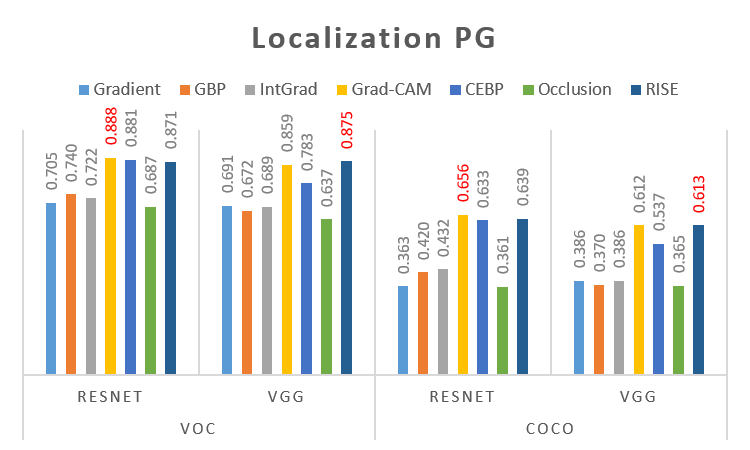}
\caption{Localization overall results with PG across VOC and COCO datasets with two pretrained models. The best performed saliency methods are highlighted. The number above every bar is the localization result based on PG. }\label{fig:localization_pg}
\end{figure}

Referring to \cite{zhang2018top}, we apply PG to quantify the localization of the XAI methods on the object to classify.  The overall results are shown in Figure. \ref{fig:localization_pg}.

From Figure. \ref{fig:localization_pg}, there are a few conclusions that can be drawn: (1) Grad-CAM and RISE get the best results. The reason behind is likely to be the same as the one analyzed in faithfulness: ResNet50 has deeper network structure; (2) CEBP performs comparably well in ResNet50 as Grad-CAM and RISE. This might be due to the fact that CEBP also propagate the excitation to intermediate convolutional layer rather than the input layer.  However for VGG16, CEBP has a significant underperformance compared to Grad-CAM.  This might be related to the fact that VGG16 has three fully connected layers and thus be too many nodes to propagate the probability; (3) The rest of the methods have marginally worse performance. As argued in Section. \ref{subsec:experiment_faithfulness}, Gradient, IntGrad, and Occlusion have noisier saliency maps. As for GBP, although the saliency map is cleaner, it highlights the edges of all objects and therefor may lose localization capability.

From the definition of PG, one can see that it measures whether an explanation indicates the ground truth bounding box. However, as it focuses only on the single point with the highest importance, it cannot well describe situations that the saliency map captures unnecessary information (saliency map of RISE in Figure. \ref{fig:faithness_saliency}). To resolve this issue, we revise slightly on the PG by taking this into account. Mathematically, we define a modified metric, that is, the ratio of Intersection between the salient area and the ground truth mask over the Salient Region (IoSR):
\begin{equation}
\text{IoSR} = \frac{M_{GT} \cap SA (E(I,f)) }{SA(E(I,f))}
\end{equation}
where $M_{GT}$ is the ground truth mask and $SA(\cdot)$ is the salient area of a saliency map. In our experiment, $SA (E(I,f))$ is calculated as the $\sum (E(I,f) >\theta \max E(I,f))$, in which  $\theta$ is a user-defined threshold (default: 0.5). 

\begin{figure}[ht!]
\centering
\includegraphics[width=1\linewidth]{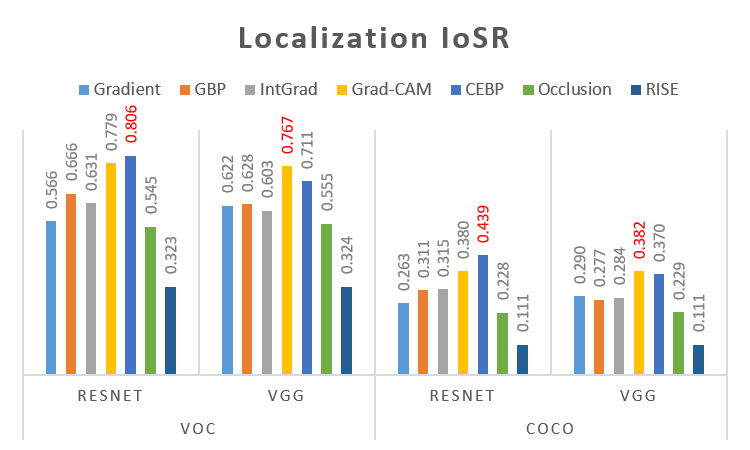}
\caption{Overall results for localization (IoSR). } \label{fig:localization_ios}
\end{figure}

Iterating through the whole set of data, the result for IoSR is shown as in Figure. \ref{fig:localization_ios}. Comparing Figure. \ref{fig:localization_pg} and Figure. \ref{fig:localization_ios}, Grad-CAM and CEBP perform comparably well in both implementations. On the contrary, RISE has the smallest score in IoSR while it gets a relatively high score in PG. The reason may lie behind the mechanisms of RISE: masking on the original images could drag the images out of data manifold, thus the resulting probability is not a mere reflection on the information the masks contain but also effects of out-of-distribution.  Occlusion is also plagued by this problem and thus Occlusion also has more significant drops on the results than the backpropagation methods. 

Although IoSR captures more information than PG, it suffers from the same issues as PG: (1) This metric is unable to distinguish the failure modes of explainer from failure modes of the pretrained model. If the result is good, it does reflect that the pretrained model can recognize the object and the explainer can locate the discriminative part of the object. However, if the result is poor, the reason could be either  the pretrained model does not understand the object or the explainer fails to find the discriminative region of the model; (2) The computation requires the human-annotated bounding boxes or segmentation masks. In most cases, this may not be available in  classification scenario.

\subsection{False-Positives Results}\label{subsec:experiment_false-postives}
In \cite{yang2019bim}, the quality of the explanations were assessed via investigating false high-score attributions on irrelevant features on BAM dataset. However, BAM dataset used in \cite{yang2019bim} only includes 10 object labels and 10 scene labels, which could lead to weak generalization for benchmarking. In this experiment, we extend BAM to eBAM by leveraging full set of objects from COCO and scenes from Miniplaces, containing 80 objects and 100 scenes. 

Following the construction of BAM, objects are extracted, rescaled to between 1/3 and 1/2 of the scene image ($128*128$ in Miniplaces), and randomly placed in the image. After extracting the objects from COCO training and validation set, a huge amount of objects are found unreasonable and defective, i.e., only tiny patches of the object are extracted, without recognizable semantic information. To relieve the impact of such defects, extracted objects are integrated into the black background and fed into Faster-RCNN (pretrained on COCO) \cite{ren2015faster} for object detection. Only the objects detected with the correct label and confidence higher than $50\%$ are retained for data synthesis. Numerous data are filtered after detection, we abandon classes containing fewer than 500 samples, remaining 53 object labels. Scenery images could also contain objects. To resolve this problem, scenery images are also inferenced through the object detection model. If any object is detected with high confidence (higher than $50\%$), the image will be discarded. Part of the images are filtered but still remain all the 100 labels. Consequently, the training set is made of 53 object labels and 100 scene labels, resulting in 5300 combined labels and each label contains nearly 500 images, a total of 2,646,328 images as the training set. We inherit the same object and scene labels on the validation set and filter the objects and scenes through the same process, leaving 378,830 images for evaluations. We train ResNet50 \cite{he2016deep} and VGG16 \cite{simonyan2014very} on the training set and evaluate MCS as well as IDR of the explaining methods on the validation set. Since perturbation-based methods take a long time on running through such a large validation set, only backpropagation methods are evaluated in this subsection.

\begin{table}[ht!]
	\caption{FP results of backpropagation methods on eBAM}
	\vspace{-3ex}
	\label{tab:FP}
	\renewcommand\arraystretch{1.2}
	\resizebox{0.47\textwidth}{2.0cm}{
	\begin{tabular}{c|c|c|c|c|c|c}
		\toprule
		&                         Model               & \textbf{Gradient} & \textbf{GBP} & \textbf{IntGrad} & \textbf{Grad-CAM} & \textbf{CEBP}   \\ \midrule
		\multirow{2}{*}{\textbf{$G_{cf}(f_s, \mathbb{I})$}} & Resnet50 & 0.019            & 0.027       & 0.029           & 0.261            & 0.025          \\ \cline{2-7} 
		& VGG16    & 0.018             & 0.009        & 0.031            & 0.114             & 0.030           \\ \hline
		\multirow{2}{*}{\textbf{$G_{cf}(f_o, \mathbb{I})$}} & Resnet50 & 0.090            & 0.050       & 0.076           & 0.754            & 0.293          \\ \cline{2-7} 
		& VGG16    & 0.105             & 0.060        & 0.079            & 0.436             & 0.261           \\ \hline
		\multirow{2}{*}{\textbf{MCS}}                       & Resnet50 & 0.071            & 0.024       & 0.046           & \textbf{0.492}   & 0.268          \\ \cline{2-7} 
		& VGG16    & 0.088             & 0.051        & 0.048            & \textbf{0.320}    & 0.231           \\ \hline
		\multirow{2}{*}{\textbf{MCR}}                       & Resnet50 & 0.310            & 0.301       & 0.285           & 0.205            & \textbf{0.653} \\ \cline{2-7} 
		& VGG16    & 0.666             & 0.630        & 0.633            & 0.693             & \textbf{0.774}  \\ \hline
		\multirow{2}{*}{\textbf{IDR}}                       & Resnet50 & \textbf{0.005}    & 0.201        & 0.112            & 0.090             & 0.203           \\ \cline{2-7} 
		& VGG16    & \textbf{0.006}    & 0.027        & 0.124            & 0.196             & 0.283           \\ \bottomrule
	\end{tabular}}
\end{table}

\subsubsection{MCS and MCR}\label{subsubsec:mcs_and_mcr}
\begin{figure}[htp]
	\centering
	\includegraphics[width = 8.7cm]{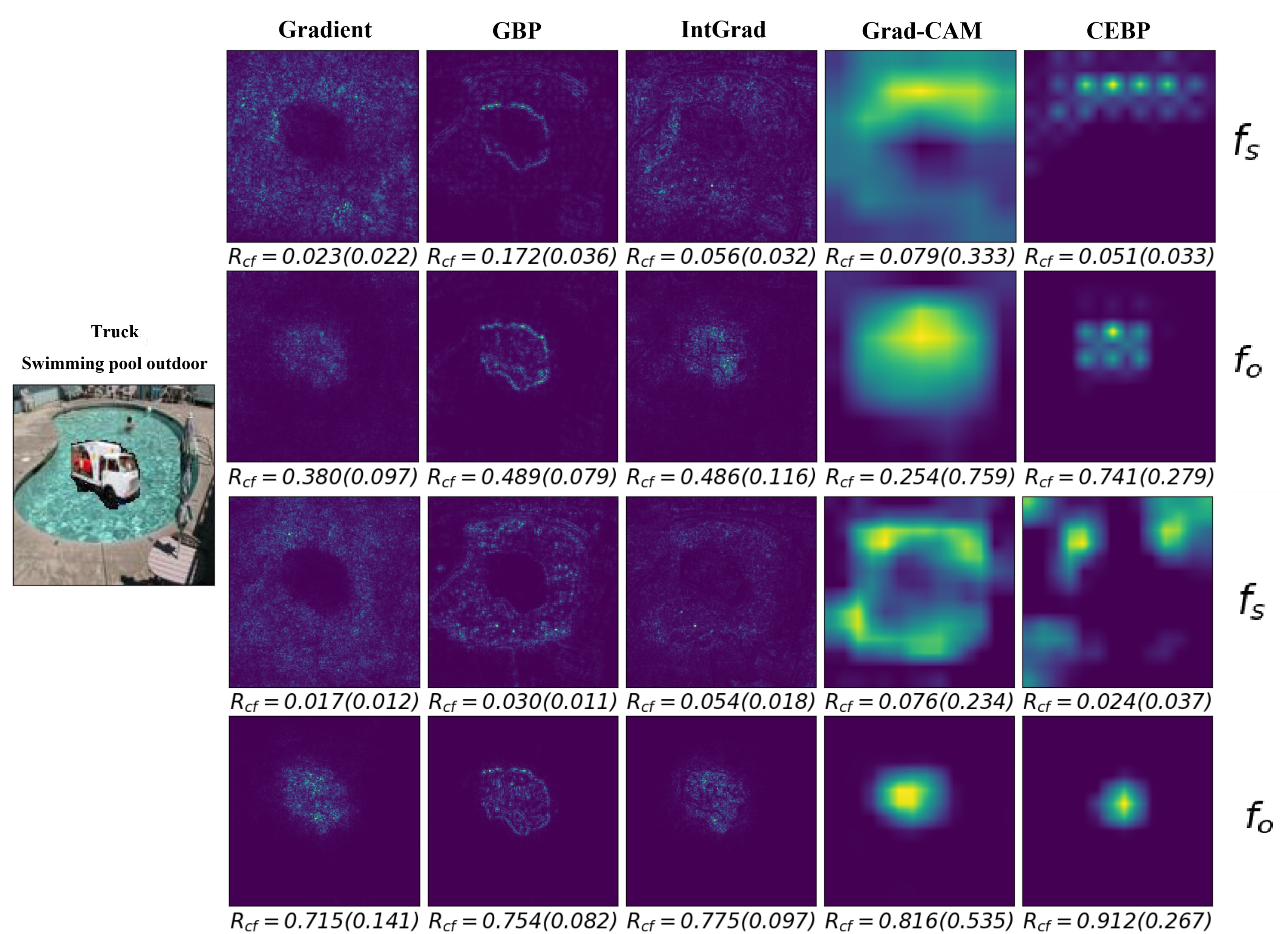}
	\caption{Saliency maps of the same synthetic images explained on object classifier $f_o$ and scene classifier $f_s$. Objects in the images are expected to be important for $f_o$ but irrelevant for $f_s$. The title above original images indicates the scene and object. The values below the saliency maps are MCR scores with MCS in parenthesis. The top and the bottom figures demonstrate the results of Resnet50 and VGG16 respectively.}
	\label{fig:mcs}
\end{figure}

$\newline$
MCS calculates the difference between concept contributions of different classifiers over the dataset. However, relevance scores in saliency maps from different interpreting methods could distribute in different ranges, even after normalization. As seen in Table.\ref{tab:FP}, Grad-CAM tends to assign a higher score than the other explanations, resulting in higher but unfair MCS. To relieve this unfairness, we further introduce Model contrast Ratio (MCR) to compare attribution differences. Given a scene classifier $f_s$, attributions on CF ought to take a low proportion of the overall saliency map. On the other hand, attributions from an object classifier $f_o$ are expected to focus on the CF region. MCR compares the ratios of CF attributions between $f_s$ and $f_o$, formulated below:

\begin{equation}\label{eq:MCR}
	MCR = \frac{1}{|\mathbb{I}_{corr}|} \sum_{I_k \in \mathbb{I}_{corr}} (\frac{S_{cf}(f_o, I_k)}{\sum (f_o, I_k)} - \frac{S_{cf}(f_o, I_k)}{\sum (f_o, I_k)})
\end{equation}
Higher MCR indicates a more accurate explanation. As shown in Table. \ref{tab:FP}, CEBP outperforms other methods in terms of MCR on both models. Grad-CAM outperforms other methods in terms of MCS, but receive the lowest MCR on Resnet50. We visualize the saliency maps of $f_s$ and $f_o$ for further checks. Demonstrated in Figure.\ref{fig:mcs}, all the backpropagation methods in the experiment exhibit weakly-supervised object localizing capability. Grad-CAM produces low-resolution saliency maps on Resnet50 as the dimension of selected convolutional layer output is $7\times 7\times 2048$, while other methods accurately highlight pixels within the CF region. High-score attributions of Grad-CAM result in higher MCS, but in the meanwhile, coarse edges pull down the MCR scores due to misleading attributions outside the objects. Due to higher resolution of selected feature maps in VGG16 ($14\times 14\times 512$), Grad-CAM generates more precise saliency maps and results in a competitive MCR score.

\subsubsection{IDR}\label{subsubsec:idr}
\begin{figure}[htp]
	\centering
	\includegraphics[width = 8.7cm]{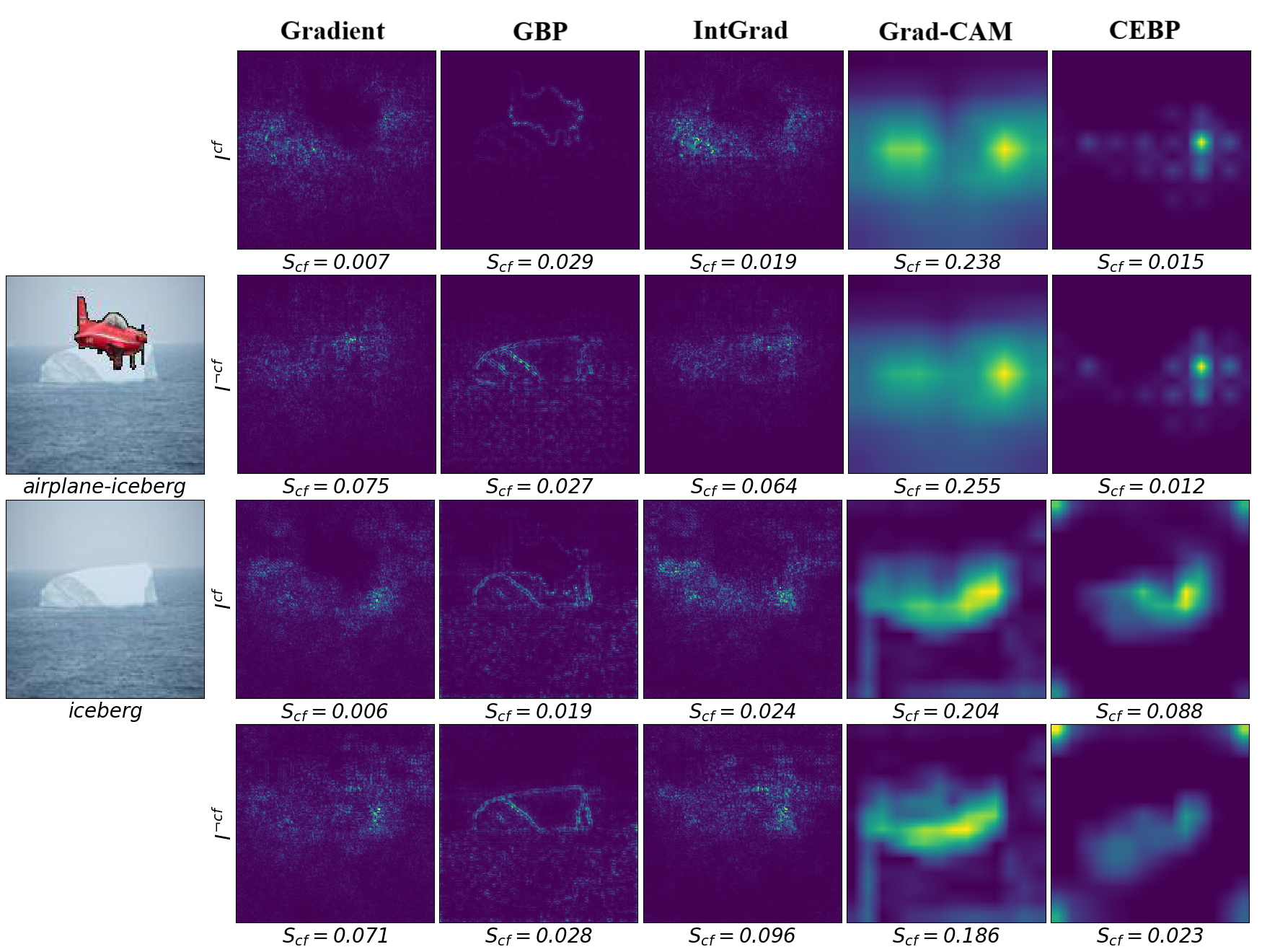}
	\caption{Saliency maps with and without CF. "Airplane" is the CF in scene classification, which should be assigned lower salient scores. The top and the bottom figures demonstrate the results of Resnet50 and VGG16 respectively.}
	\label{fig:fpr1}
\end{figure}

$\newline$
Higher IDR indicates that the saliency method assigns higher scores to CF regions which are considered irrelevant in decision making. The baseline of IDR is 50\%, representing the performance of randomly attributed saliency maps. As shown in Table. \ref{tab:FP}, CEBP exhibits unreasonable attributions on both Resnet50 and VGG16 while Gradient performs the best in IDR test (consistent with the conclusion of \cite{yang2019bim}). Grad-CAM gets the opposite ranking on Resnet50 and VGG16. Figure. \ref{fig:fpr1} demonstrates the saliency maps of images with and without CF. In the example of "airplane-iceberg", Gradient and IntGrad have clear dark spots on the CF, exhibiting better discriminating ability. We notice that GBP assigns a high relevance score to the edge of CF object, as suggested in \cite{adebayo2018sanity}.

We measure false-positives of explanations on eBAM dataset in this section. To relieve the unfairness introduced by MCS, we propose MCR to compare attributions from different classifiers. Drawing the same conclusion as \cite{yang2019bim}, Gradient performs well in both MCR and IDR measurements, although it only processes a simple backpropagation to produce saliency maps. A different finding is that Grad-CAM falls back on IDR in the experiment on Resnet50, while it is reported as good as Gradient on BAM dataset. This situation indicates that different datasets could lead to different FP results. We also observe different rankings on the same model structure and metrics but with various hyper-parameters. Therefore, generalizing the explanation ranking from eBAM or BAM to other datasets could be questionable. On the other hand, such benchmark could still be considered as a rigorous test to develop new explaining methods.

\subsection{Sensitivity Check Results} 
\begin{figure}[htp]
	\centering
	\includegraphics[width = 8.3cm]{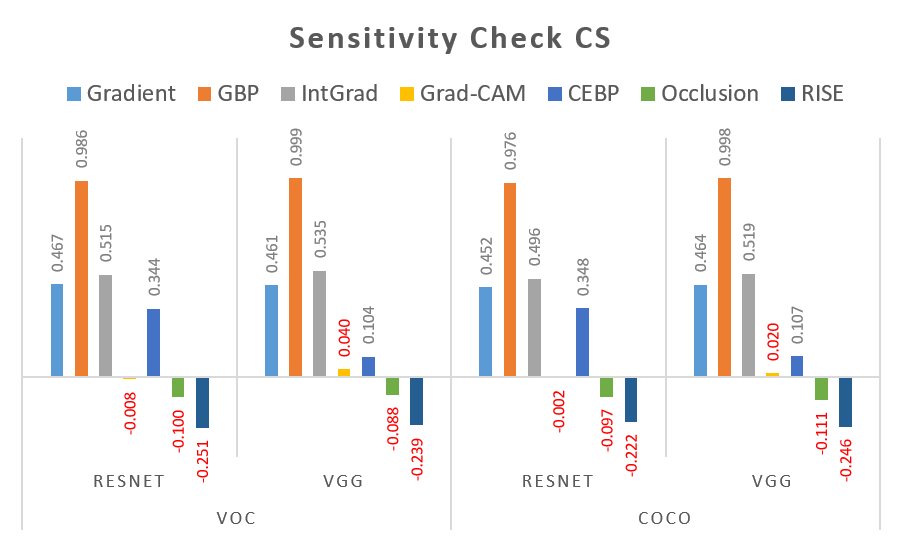}
	\caption{Sensitivity check results with CS (calculated by Pearson correlation) across two datasets and two pretrained models. Well-performed saliency methods that have negative  are highlighted.}
	\label{fig:class_sensitivity_results}
\end{figure}

For the simplicity of the experiments, we choose to measure the class sensitivity of the XAI methods.  Following the definition of class sensitivity described in \ref{subsec:sensitivity_check}, we measure the similarity between max-confidence and min-confidence saliency maps.  In this work, we apply Pearson correlation to perform experiments for CS. Specifically, smaller correlation, i.e., larger discrepancy, between maximal and minimal confidence saliency map of the original image indicates higher class sensitivity of the explaining method. Figure. \ref{fig:class_sensitivity_results} shows the comparison among CS of the selected saliency methods applied across two datasets and two pretrained models. 

\begin{figure*}[htp]
	\centering
	\includegraphics[width = 16cm]{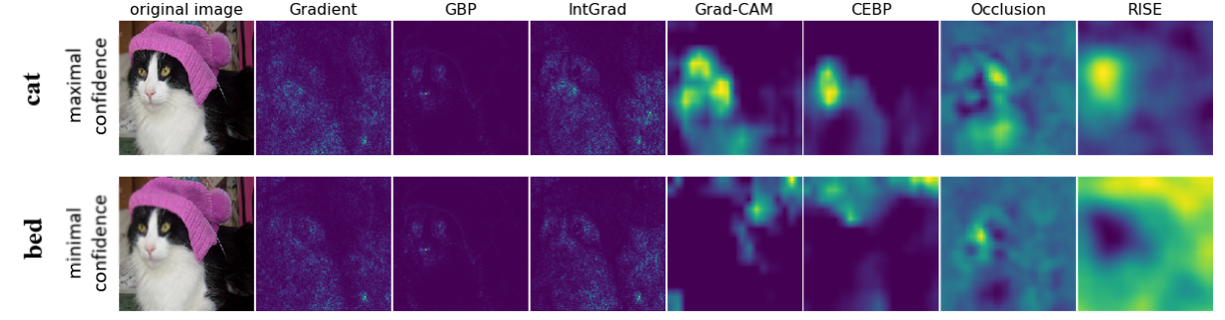}
	\caption{Typical maximal (1st row) and minimal (2nd row) confidence saliency maps of the listed methods. Data samples are from COCO and the pretrained model is VGG16.}
	\label{fig:class_sensitivity_example}
\end{figure*}

From the overall experimental results in Figure. \ref{fig:class_sensitivity_results}, we can see that: Grad-CAM, RISE, and Occlusion perform fairly well w.r.t class sensitivity as the CS scores are nearly zero or below zero. GBP has the worst class sensitivity. For the rest of the methods, the ranking is CEBP > Gradient > IntGrad. Figure. \ref{fig:class_sensitivity_example} demonstrates an example from COCO dataset and the visual comparison supports the ranking above. The saliency maps of max-confidence label "cat" and min-confidence label "bed" are almost identical from GBP. As explained by \cite{adebayo2018sanity}, GBP acts like an edge detector, and thus results in poor class sensitivity. Another interesting result suggested in Figure. \ref{fig:class_sensitivity_results} is that perturbation-based methods have negative class sensitivity. As shown in Figure. \ref{fig:class_sensitivity_example}, the saliency maps of max-confidence and min-confidence labels highlight complementary regions.

\subsection{Stability Results}
\begin{figure}[!h]
\centering
\includegraphics[width=0.8\columnwidth]{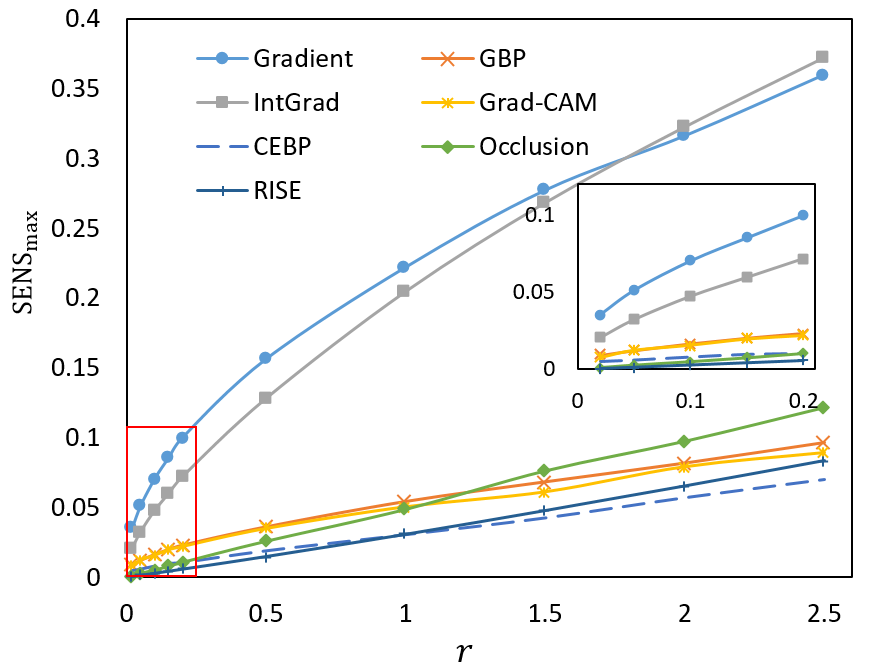}
\caption{$\mathrm{SENS_{max}}$ vs. perturbation radius $r$ relation. The $x$-axis is the settings of perturbation radius during experiments and the $y$-axis is the calculated results for $\mathrm{SENS_{max}}$. The inset is the zoom-in view of the rectangular framed region. Different curves represent different saliency methods behavior. The results are obtained from averaging five examples from PASCAL VOC and the pretrained model is VGG16.}\label{fig:sens_vs_r}
\end{figure}

Following the definition of $\mathrm{SENS_{max}}$ reviewed in Section. \ref{subsec:Robustness}, we quantitatively measure the stability for the saliency methods mentioned in Section. \ref{sec:method}. One might notice that $\mathrm{SENS_{max}}$ depends on the radius of perturbation. To investigate such dependence, we first varied the radius and perturbation and get $\mathrm{SENS_{max}}$ on few examples (the results are averages across five examples) and get the results shown in Figure. \ref{fig:sens_vs_r}.

As demonstrated in Figure. \ref{fig:sens_vs_r}, metric $\mathrm{SENS_{max}}$ strongly depends on the choice of radius of perturbation. This indicates that stability of a saliency method varies w.r.t the strength of perturbation. In the cases of small perturbation radius, RISE and Occlusion get smaller sensitivity thus show higher stability while Gradient exhibits high sensitivity. However, when perturbation becomes stronger, the curves start to cross with each other. In the stronger perturbation region, CEBP and RISE exhibit the best stability. In real life applications, the stability of saliency methods should be measured within the range of the practical perturbation.

\begin{figure}
\centering
\includegraphics[width=1\columnwidth]{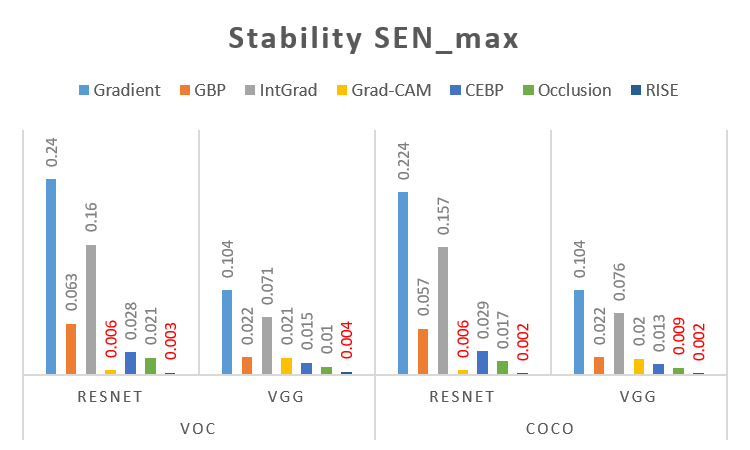}
\caption{Stability results from $\mathrm{SENS_{max}}$ in small perturbation region with perturbation radius $r=0.2$ across two different datasets and two pretrained models. The results are averaged across 50 randomly selected examples from both datasets.}\label{fig:stability_result}
\end{figure}

Considering a small perturbation region (as the definition of stability is on the insignificant perturbation), the $\mathrm{SENS_{max}}$ results across different datasets and different pretrained models are obtained in Figure. \ref{fig:stability_result}. 
Figure \ref{fig:stability_result} shows that Gradient and IntGrad have the marginally worse stability compared to others. The possible reason for this is that these two methods contains negative gradients and the perturbation noise most likely has negative contribution to the final output. Other methods would not suffer from this issue. 
RISE and Grad-CAM exhibit the strongest stability in Resnet50 while RISE and Occlusion behave the best for VGG16. The stability of Grad-CAM may also be related to the fact that it calculates the attribution of last convolutional layer. As for RISE and Occlusion, the calculation process already involve perturbation and thus a slight perturbation on the input may not have significant effects. Another finding from Figure. \ref{fig:stability_result} is that this quantity does not show significant variations w.r.t to datasets: not only the relative rankings are the same for both datasets, the values for different methods are also nearly identical across the two datasets. 

\subsection{Summary of Experimental Results}
In this subsection, we summarize the above experimental results. To better show these results, we average the four experimental results across the two datasets and two pretrained models of faithfulness, localization, sensitivity check, and stability. We then get the results in Table. \ref{tab:average_results}, with the marginally better methods highlighted for every metric. In terms of faithfulness, Grad-CAM and RISE perform the best. As for sensitivity check, Grad-CAM, RISE, and Occlusion have excellent performance while GBP has almost no class sensitivity. And RISE has the best stability while Gradient seems to be sensitive to insignificant perturbation. From Table. \ref{tab:FP}, Gradient exhibits solid performance in the false-positives experiment. Advanced methods derived from Gradient obtain lower IDR. According to the results above, we find no saliency method could serve as a silver bullet in saliency-based problems, explaining methods need carefully selected depending on the desired properties.

\begin{table}[h!]
\caption{Averaged Experimental Results Across Two Datasets and Two Pretrained Models}
\scalebox{0.76}{
\begin{tabular}{l|l|l|l|l|l}
\toprule
 XAI methods         & \multicolumn{1}{c|}{\begin{tabular}[c]{@{}c@{}}Localization\\      (PG)\end{tabular}} & \multicolumn{1}{c|}{\begin{tabular}[c]{@{}c@{}}Localization\\      (IoSR)\end{tabular}} & \multicolumn{1}{c|}{\begin{tabular}[c]{@{}c@{}}Faithfulness\\      (iAUC)\end{tabular}} & \multicolumn{1}{c|}{\begin{tabular}[c]{@{}c@{}}Sensitivity\\      Check (CS)\end{tabular}} & Stability      \\ \midrule
Gradient  & 0.536                                                                                 & 0.435                                                                                   & 0.282                                                                                   & 0.461                                                                                      & 0.168          \\ \hline
GBP       & 0.551                                                                                 & 0.47                                                                                    & 0.384                                                                                   & 0.99                                                                                       & 0.041          \\ \hline
IntGrad   & 0.557                                                                                 & 0.458                                                                                   & 0.286                                                                                   & 0.516                                                                                      & 0.116          \\ \hline
Grad-CAM  & \textbf{0.754}                                                                        & \textbf{0.577}                                                                          & \textbf{0.43}                                                                           & \textbf{0.013}                                                                             & 0.013          \\ \hline
CEBP      & 0.709                                                                                 & \textbf{0.582}                                                                          & 0.397                                                                                   & 0.226                                                                                      & 0.021          \\ \hline
Occlusion & 0.512                                                                                 & 0.389                                                                                   & 0.308                                                                                   & \textbf{-0.099}                                                                            & 0.014          \\ \hline
RISE      & \textbf{0.75}                                                                         & 0.217                                                                                   & \textbf{0.434}                                                                          & \textbf{-0.239}                                                                            & \textbf{0.002} \\ \bottomrule
\end{tabular}\label{tab:average_results}}
\end{table}


\section{Utilization of XAI Methods and Metrics}\label{sec:utilization}
In the previous section, we mainly discuss the overall performance of the XAI methods with the metrics reviewed in Section \ref{sec:metric}. In this section, we present a case study of the XAI methods as well as its metrics in actual model analysis -- \textit{Clever Hans Detection}. "Clever Hans" originally refers to a horse that was believed to be able to perform arithmetic calculations. Later it was discovered that instead of conducting actual calculation, this horse was giving the correct answers based on the unintended gestures of its trainer. Similar Clever Hans phenomena also occur in machine learning, that the trained machine make decisions based on spurious correlations, e.g., the trained Fisher Vector classifier classifies an image to be "horse" based on the watermark \cite{lapuschkin2019unmasking}.  
A model that learned "Clever Hans"-type decision strategy will likely fail to provide correct classification for new datasets without such spurious correlations. Thus it is essential to find out the Clever Hans cases of the model before actual deployment. However, conducting visual inspections one by one on the explanations of whole dataset can be laborious costive and most likely unable to execute. In this section, we propose a simple way to find out the examples that show Clever Hans effects in a classification scenario with XAI methods and measuring metrics.

Revisiting the concept of Clever Hans, such examples would likely show high classification confidence but the classification basis for them are "wrong". A faithful explanation help to locate such classification basis. Thus, borrowing the metrics studied in the above sections, we can have a simple formula to find out the Clever Hans examples $\{I_{CH}\}$ from a large dataset:
\begin{equation}\label{eq:clever_hans}
\{I_{CH}\} = \{I_{f_c>\theta_c}\}\cap \{I_{M_{loc} <\theta_l}\} \cap \{I_{M_{faith} >\theta_f}\}
\end{equation}
where $\{I_{f_c>\theta_c}\}$ refers to the images set with model output $f_c$ greater than a threshold $\theta_c$, and  $\{I_{M_{loc}<\theta_l}\}$ and $\{I_{M_{faith}>\theta_f}\}$ refer to the image set with metric of localization lower than a threshold $\theta_l$ and metric of faithfulness greater that a threshold $\theta_f$, respectively. Notably, the metric of faithfulness is applied here to make sure the explanation is faithful enough.
   
\begin{figure}
\includegraphics[width=0.9\linewidth]{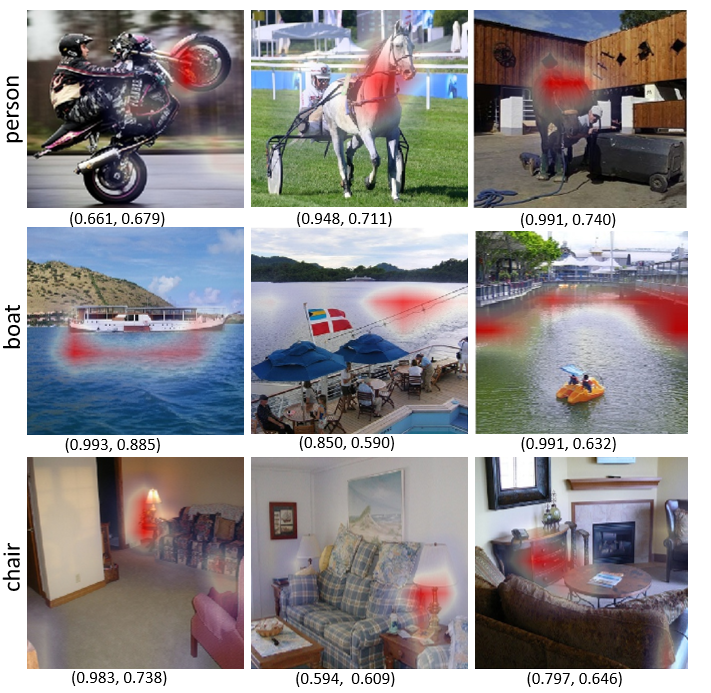}
\caption{Selected examples from VOC dataset that  show Clever Hans phenomenon. The predicted labels for rows from top to bottom are "person", "aeroplane", and "chair", respectively. The predicted outputs and the iAUC metric results are listed below each figure and the PG results for all these examples are 0. The explanation method we apply here is Grad-CAM and the black-box model is ResNet50.}\label{fig:cleverhans}
\end{figure}

\begin{figure}
\includegraphics[width=0.9\linewidth]{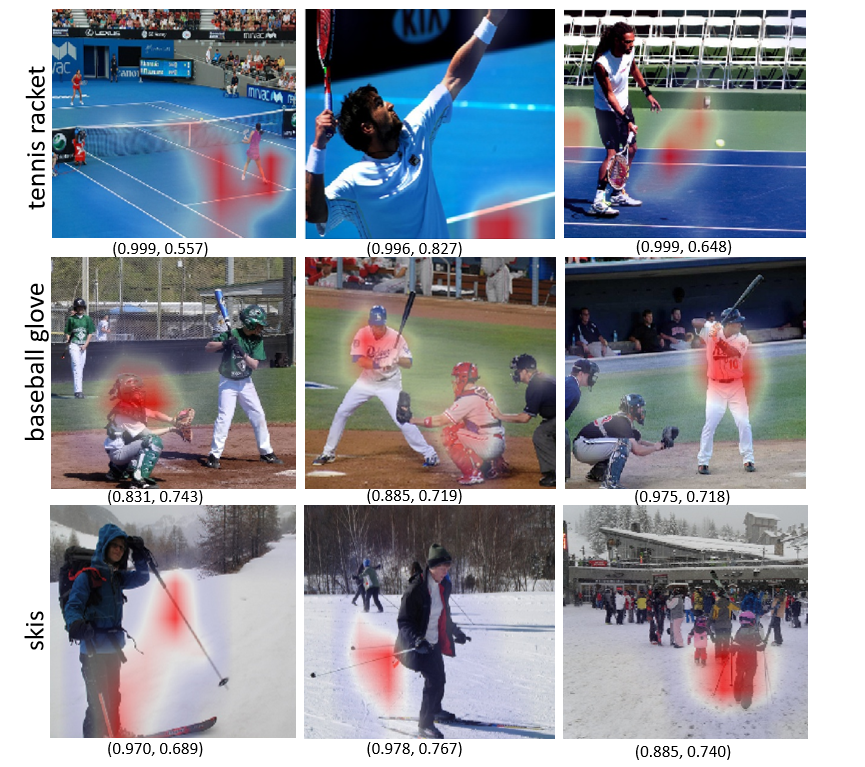}
\caption{Selected examples from COCO dataset that show Clever Hans phenomenon. The predicted outputs and the iAUC metric results are listed below each figure and the PG results for all these examples are 0. The classification model here is ResNet50 for COCO.}\label{fig:cleverhans_coco}
\end{figure}

Based on Equation. (\ref{eq:clever_hans}), we find a set of examples that show Clever Hans phenomenon. Figure. \ref{fig:cleverhans} shows a few representative examples of them. In these examples, although the black-box model gives correct predictions, the found explainable classification bases are unreasonable. For instance, although the model correctly recognizes "person", its bases are the presence of "horse" or "motorbike" as shown in the first row in Figure. \ref{fig:cleverhans}. This may be caused by the concurrence of "person" and "horse" or "motorbike" in the training set. Other selected interesting examples from COCO are shown in Figure. \ref{fig:cleverhans_coco}. The model predicts "tennis racket", "baseball glove", and "skis" based on tennis court, the outfit of the baseball player, and the skis poles, respectively.

\begin{figure}
\includegraphics[width=0.9\linewidth]{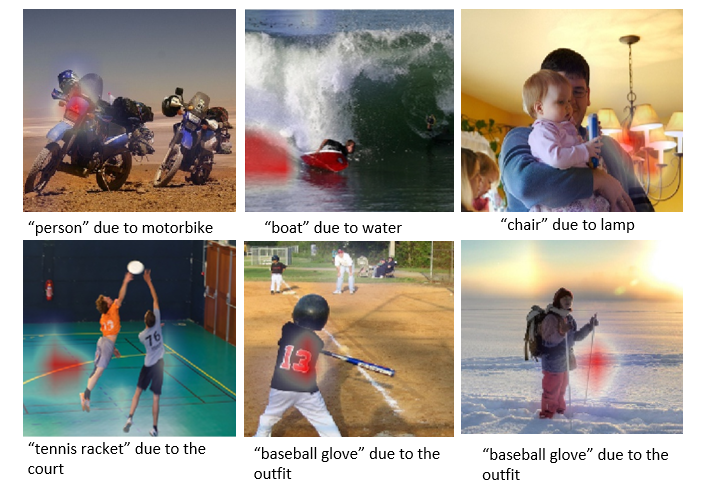}
\caption{False-positive examples due to "Clever Hans"-type features. }\label{fig:false_postives}
\end{figure}

As we mentioned that "Clever Hans"-type classification strategies may cause misclassification when dealing with data without spurious correlations. As an investigation of this effect, we check on the failure modes on the test dataset. We list a few examples in Figure. \ref{fig:false_postives}, where the black-box model makes false-positive predictions due to the presence of "Clever Hans"-type features. An image containing only "motorbike" is predicted to be positive about the presence of "person" because the model takes the presence of "motorbike" feature as a classification base for "person". 



\section{Conclusions and Outlook}\label{sec:conclusion}
In conclusion, we first filter a set of metrics that are proposed to quantify the quality of XAI methods from five different perspectives. Then, we conduct a thorough set of experiments to quantitatively evaluate some of the popular saliency methods with these metrics. The experimental results show that no single method can achieve the best results from all perspectives. In application, users need to choose an explanation method according to their priorities. Nevertheless, Grad-CAM and RISE perform quite well for all metrics except false-positives for Grad-CAM and IoSR of localization for RISE. This is due to their intrinsic natures: the resolution of Grad-CAM is very low compared to other backpropagation methods, and as a perturbation method RISE usually has dispersive saliency maps. 

Along with the experiments, we propose a new metric -- Intersection between the salient region and the bounding box over Salient Region (IoSR) for localization to capture whether the saliency map is compact. And to calculate the false-positives, we expand the original BAM dataset to accommodate more objects and more scenes. Last but not least, we propose a simple application of a combination of metrics to locate Clever Hans examples and analyze the failure modes of the black-box models. 

Our last comment is that, although the community may expect a universal metric set to evaluate all explanation methods, nature and applying scenario of XAI might turn us down. The ultimate purpose of XAI, arguably, is to provide necessary information to satisfy the causal curiosity of the user to build trust. To this end, fair evaluation can only be effectively conducted under a fixed scenario and mode setting. Moreover, since the explanation is for human users to build their understanding and trust on the machine side, a sound evaluation metric should reside deeply on cognitive rationales, and currently, there is an observable gap between the cognitive findings and their computational implementations. We believe continuous research attention in this direction will finally bring about insightful results. As for the applications of XAI, we also note here finding Clever Hans examples in the datasets as  shown in Section. \ref{sec:utilization} can not only help diagnose the black-box models, but might also be applied to further improve the generalization and the performance of the model through simple data modification, e.g. masking the salient part of the examples that show Clever Hans phenomena, and retrain the black-box model.

\balance
\bibliographystyle{ACM-Reference-Format}
\bibliography{ref}


\begin{thebibliography}{43}


\ifx \showCODEN    \undefined \def \showCODEN     #1{\unskip}     \fi
\ifx \showDOI      \undefined \def \showDOI       #1{#1}\fi
\ifx \showISBNx    \undefined \def \showISBNx     #1{\unskip}     \fi
\ifx \showISBNxiii \undefined \def \showISBNxiii  #1{\unskip}     \fi
\ifx \showISSN     \undefined \def \showISSN      #1{\unskip}     \fi
\ifx \showLCCN     \undefined \def \showLCCN      #1{\unskip}     \fi
\ifx \shownote     \undefined \def \shownote      #1{#1}          \fi
\ifx \showarticletitle \undefined \def \showarticletitle #1{#1}   \fi
\ifx \showURL      \undefined \def \showURL       {\relax}        \fi
\providecommand\bibfield[2]{#2}
\providecommand\bibinfo[2]{#2}
\providecommand\natexlab[1]{#1}
\providecommand\showeprint[2][]{arXiv:#2}

\bibitem[\protect\citeauthoryear{Adadi and Berrada}{Adadi and Berrada}{2018}]%
        {adadi2018peeking}
\bibfield{author}{\bibinfo{person}{Amina Adadi} {and} \bibinfo{person}{Mohammed
  Berrada}.} \bibinfo{year}{2018}\natexlab{}.
\newblock \showarticletitle{Peeking inside the black-box: A survey on
  Explainable Artificial Intelligence (XAI)}.
\newblock \bibinfo{journal}{\emph{IEEE Access}}  \bibinfo{volume}{6}
  (\bibinfo{year}{2018}), \bibinfo{pages}{52138--52160}.
\newblock


\bibitem[\protect\citeauthoryear{Adebayo, Gilmer, Muelly, Goodfellow, Hardt,
  and Kim}{Adebayo et~al\mbox{.}}{2018}]%
        {adebayo2018sanity}
\bibfield{author}{\bibinfo{person}{Julius Adebayo}, \bibinfo{person}{Justin
  Gilmer}, \bibinfo{person}{Michael Muelly}, \bibinfo{person}{Ian Goodfellow},
  \bibinfo{person}{Moritz Hardt}, {and} \bibinfo{person}{Been Kim}.}
  \bibinfo{year}{2018}\natexlab{}.
\newblock \showarticletitle{Sanity checks for saliency maps}. In
  \bibinfo{booktitle}{\emph{Advances in Neural Information Processing
  Systems}}. \bibinfo{pages}{9505--9515}.
\newblock


\bibitem[\protect\citeauthoryear{Boer, Deutch, Frost, and Milo}{Boer
  et~al\mbox{.}}{2020}]%
        {boer2020personal}
\bibfield{author}{\bibinfo{person}{Naama Boer}, \bibinfo{person}{Daniel
  Deutch}, \bibinfo{person}{Nave Frost}, {and} \bibinfo{person}{Tova Milo}.}
  \bibinfo{year}{2020}\natexlab{}.
\newblock \showarticletitle{Personal insights for altering decisions of
  tree-based ensembles over time}.
\newblock \bibinfo{journal}{\emph{Proceedings of the VLDB Endowment}}
  \bibinfo{volume}{13}, \bibinfo{number}{6} (\bibinfo{year}{2020}),
  \bibinfo{pages}{798--811}.
\newblock


\bibitem[\protect\citeauthoryear{Dombrowski, Alber, Anders, Ackermann,
  M{\"u}ller, and Kessel}{Dombrowski et~al\mbox{.}}{2019}]%
        {dombrowski2019explanations}
\bibfield{author}{\bibinfo{person}{Ann-Kathrin Dombrowski},
  \bibinfo{person}{Maximillian Alber}, \bibinfo{person}{Christopher Anders},
  \bibinfo{person}{Marcel Ackermann}, \bibinfo{person}{Klaus-Robert
  M{\"u}ller}, {and} \bibinfo{person}{Pan Kessel}.}
  \bibinfo{year}{2019}\natexlab{}.
\newblock \showarticletitle{Explanations can be manipulated and geometry is to
  blame}. In \bibinfo{booktitle}{\emph{Advances in Neural Information
  Processing Systems}}. \bibinfo{pages}{13567--13578}.
\newblock


\bibitem[\protect\citeauthoryear{Doshi-Velez and Kim}{Doshi-Velez and
  Kim}{2017}]%
        {doshi2017towards}
\bibfield{author}{\bibinfo{person}{Finale Doshi-Velez} {and}
  \bibinfo{person}{Been Kim}.} \bibinfo{year}{2017}\natexlab{}.
\newblock \showarticletitle{Towards a rigorous science of interpretable machine
  learning}.
\newblock \bibinfo{journal}{\emph{arXiv preprint arXiv:1702.08608}}
  (\bibinfo{year}{2017}).
\newblock


\bibitem[\protect\citeauthoryear{Everingham, Van~Gool, Williams, Winn, and
  Zisserman}{Everingham et~al\mbox{.}}{[n. d.]}]%
        {pascal-voc-2007}
\bibfield{author}{\bibinfo{person}{M. Everingham}, \bibinfo{person}{L.
  Van~Gool}, \bibinfo{person}{C.~K.~I. Williams}, \bibinfo{person}{J. Winn},
  {and} \bibinfo{person}{A. Zisserman}.} \bibinfo{year}{[n. d.]}\natexlab{}.
\newblock \bibinfo{title}{The {PASCAL} {V}isual {O}bject {C}lasses {C}hallenge
  2007 {(VOC2007)} {R}esults}.
\newblock
  \bibinfo{howpublished}{http://www.pascal-network.org/challenges/VOC/voc2007/workshop/index.html}.
\newblock


\bibitem[\protect\citeauthoryear{Guillaumin and Ferrari}{Guillaumin and
  Ferrari}{2012}]%
        {guillaumin2012large}
\bibfield{author}{\bibinfo{person}{Matthieu Guillaumin} {and}
  \bibinfo{person}{Vittorio Ferrari}.} \bibinfo{year}{2012}\natexlab{}.
\newblock \showarticletitle{Large-scale knowledge transfer for object
  localization in imagenet}. In \bibinfo{booktitle}{\emph{2012 IEEE Conference
  on Computer Vision and Pattern Recognition}}. IEEE,
  \bibinfo{pages}{3202--3209}.
\newblock


\bibitem[\protect\citeauthoryear{He, Zhang, Ren, and Sun}{He
  et~al\mbox{.}}{2016}]%
        {he2016deep}
\bibfield{author}{\bibinfo{person}{Kaiming He}, \bibinfo{person}{Xiangyu
  Zhang}, \bibinfo{person}{Shaoqing Ren}, {and} \bibinfo{person}{Jian Sun}.}
  \bibinfo{year}{2016}\natexlab{}.
\newblock \showarticletitle{Deep residual learning for image recognition}. In
  \bibinfo{booktitle}{\emph{Proceedings of the IEEE conference on computer
  vision and pattern recognition}}. \bibinfo{pages}{770--778}.
\newblock


\bibitem[\protect\citeauthoryear{Hoffman, Mueller, Klein, and Litman}{Hoffman
  et~al\mbox{.}}{2018}]%
        {hoffman2018metrics}
\bibfield{author}{\bibinfo{person}{Robert~R. Hoffman},
  \bibinfo{person}{Shane~T. Mueller}, \bibinfo{person}{Gary Klein}, {and}
  \bibinfo{person}{Jordan Litman}.} \bibinfo{year}{2018}\natexlab{}.
\newblock \showarticletitle{Metrics for explainable {AI}: Challenges and
  prospects}.
\newblock \bibinfo{journal}{\emph{arXiv preprint arXiv:1812.04608}}
  (\bibinfo{year}{2018}).
\newblock


\bibitem[\protect\citeauthoryear{Kim, Wattenberg, Gilmer, Cai, Wexler, Viegas,
  and Sayres}{Kim et~al\mbox{.}}{2017}]%
        {kim2017interpretability}
\bibfield{author}{\bibinfo{person}{Been Kim}, \bibinfo{person}{Martin
  Wattenberg}, \bibinfo{person}{Justin Gilmer}, \bibinfo{person}{Carrie Cai},
  \bibinfo{person}{James Wexler}, \bibinfo{person}{Fernanda Viegas}, {and}
  \bibinfo{person}{Rory Sayres}.} \bibinfo{year}{2017}\natexlab{}.
\newblock \showarticletitle{Interpretability beyond feature attribution:
  Quantitative testing with concept activation vectors (tcav)}.
\newblock \bibinfo{journal}{\emph{arXiv preprint arXiv:1711.11279}}
  (\bibinfo{year}{2017}).
\newblock


\bibitem[\protect\citeauthoryear{Krishnan and Wu}{Krishnan and Wu}{2017}]%
        {krishnan2017palm}
\bibfield{author}{\bibinfo{person}{Sanjay Krishnan} {and}
  \bibinfo{person}{Eugene Wu}.} \bibinfo{year}{2017}\natexlab{}.
\newblock \showarticletitle{Palm: Machine learning explanations for iterative
  debugging}. In \bibinfo{booktitle}{\emph{Proceedings of the 2nd Workshop on
  Human-In-the-Loop Data Analytics}}. \bibinfo{pages}{1--6}.
\newblock


\bibitem[\protect\citeauthoryear{Krizhevsky, Sutskever, and Hinton}{Krizhevsky
  et~al\mbox{.}}{2012}]%
        {krizhevsky2012imagenet}
\bibfield{author}{\bibinfo{person}{Alex Krizhevsky}, \bibinfo{person}{Ilya
  Sutskever}, {and} \bibinfo{person}{Geoffrey~E. Hinton}.}
  \bibinfo{year}{2012}\natexlab{}.
\newblock \showarticletitle{Imagenet classification with deep convolutional
  neural networks}. In \bibinfo{booktitle}{\emph{Advances in {Neural
  Information Processing Systems}}}. \bibinfo{pages}{1097--1105}.
\newblock


\bibitem[\protect\citeauthoryear{Lapuschkin, W{\"a}ldchen, Binder, Montavon,
  Samek, and M{\"u}ller}{Lapuschkin et~al\mbox{.}}{2019}]%
        {lapuschkin2019unmasking}
\bibfield{author}{\bibinfo{person}{Sebastian Lapuschkin},
  \bibinfo{person}{Stephan W{\"a}ldchen}, \bibinfo{person}{Alexander Binder},
  \bibinfo{person}{Gr{\'e}goire Montavon}, \bibinfo{person}{Wojciech Samek},
  {and} \bibinfo{person}{Klaus-Robert M{\"u}ller}.}
  \bibinfo{year}{2019}\natexlab{}.
\newblock \showarticletitle{Unmasking clever hans predictors and assessing what
  machines really learn}.
\newblock \bibinfo{journal}{\emph{Nature communications}} \bibinfo{volume}{10},
  \bibinfo{number}{1} (\bibinfo{year}{2019}), \bibinfo{pages}{1--8}.
\newblock


\bibitem[\protect\citeauthoryear{Li, Cao, Shi, Bai, Gao, Qiu, Wang, Gao, Zhang,
  Xue, and Chen}{Li et~al\mbox{.}}{2020}]%
        {li2020survey}
\bibfield{author}{\bibinfo{person}{Xiao-Hui Li}, \bibinfo{person}{Caleb~Chen
  Cao}, \bibinfo{person}{Yuhan Shi}, \bibinfo{person}{Wei Bai},
  \bibinfo{person}{Han Gao}, \bibinfo{person}{Luyu Qiu}, \bibinfo{person}{Cong
  Wang}, \bibinfo{person}{Yuanyuan Gao}, \bibinfo{person}{Shenjia Zhang},
  \bibinfo{person}{Xun Xue}, {and} \bibinfo{person}{Lei Chen}.}
  \bibinfo{year}{2020}\natexlab{}.
\newblock \showarticletitle{A Survey of Data-driven and Knowledge-aware
  eXplainable AI}.
\newblock \bibinfo{journal}{\emph{IEEE Transactions on Knowledge and Data
  Engineering}} (\bibinfo{year}{2020}).
\newblock


\bibitem[\protect\citeauthoryear{Lin, Maire, Belongie, Hays, Perona, Ramanan,
  Doll{\'a}r, and Zitnick}{Lin et~al\mbox{.}}{2014}]%
        {lin2014microsoft}
\bibfield{author}{\bibinfo{person}{Tsung-Yi Lin}, \bibinfo{person}{Michael
  Maire}, \bibinfo{person}{Serge Belongie}, \bibinfo{person}{James Hays},
  \bibinfo{person}{Pietro Perona}, \bibinfo{person}{Deva Ramanan},
  \bibinfo{person}{Piotr Doll{\'a}r}, {and} \bibinfo{person}{C~Lawrence
  Zitnick}.} \bibinfo{year}{2014}\natexlab{}.
\newblock \showarticletitle{Microsoft coco: Common objects in context}. In
  \bibinfo{booktitle}{\emph{European conference on computer vision}}. Springer,
  \bibinfo{pages}{740--755}.
\newblock


\bibitem[\protect\citeauthoryear{Lundberg and Lee}{Lundberg and Lee}{2017}]%
        {lundberg_unified_2017}
\bibfield{author}{\bibinfo{person}{Scott~M. Lundberg} {and}
  \bibinfo{person}{Su-In Lee}.} \bibinfo{year}{2017}\natexlab{}.
\newblock \showarticletitle{A {{Unified Approach}} to {{Interpreting Model
  Predictions}}}.
\newblock In \bibinfo{booktitle}{\emph{Advances in {{Neural Information
  Processing Systems}} 30}}. \bibinfo{publisher}{{Curran Associates, Inc.}},
  \bibinfo{pages}{4765--4774}.
\newblock


\bibitem[\protect\citeauthoryear{Ma, Liu, Lee, Zhang, and Grama}{Ma
  et~al\mbox{.}}{2018}]%
        {ma2018mode}
\bibfield{author}{\bibinfo{person}{Shiqing Ma}, \bibinfo{person}{Yingqi Liu},
  \bibinfo{person}{Wen-Chuan Lee}, \bibinfo{person}{Xiangyu Zhang}, {and}
  \bibinfo{person}{Ananth Grama}.} \bibinfo{year}{2018}\natexlab{}.
\newblock \showarticletitle{MODE: automated neural network model debugging via
  state differential analysis and input selection}. In
  \bibinfo{booktitle}{\emph{Proceedings of the 2018 26th ACM Joint Meeting on
  European Software Engineering Conference and Symposium on the Foundations of
  Software Engineering}}. \bibinfo{pages}{175--186}.
\newblock


\bibitem[\protect\citeauthoryear{Melis and Jaakkola}{Melis and
  Jaakkola}{2018}]%
        {melis2018towards}
\bibfield{author}{\bibinfo{person}{David~Alvarez Melis} {and}
  \bibinfo{person}{Tommi Jaakkola}.} \bibinfo{year}{2018}\natexlab{}.
\newblock \showarticletitle{Towards robust interpretability with
  self-explaining neural networks}. In \bibinfo{booktitle}{\emph{Advances in
  Neural Information Processing Systems}}. \bibinfo{pages}{7775--7784}.
\newblock


\bibitem[\protect\citeauthoryear{Nguyen, Dosovitskiy, Yosinski, Brox, and
  Clune}{Nguyen et~al\mbox{.}}{2016}]%
        {nguyen2016synthesizing}
\bibfield{author}{\bibinfo{person}{Anh Nguyen}, \bibinfo{person}{Alexey
  Dosovitskiy}, \bibinfo{person}{Jason Yosinski}, \bibinfo{person}{Thomas
  Brox}, {and} \bibinfo{person}{Jeff Clune}.} \bibinfo{year}{2016}\natexlab{}.
\newblock \showarticletitle{Synthesizing the preferred inputs for neurons in
  neural networks via deep generator networks}. In
  \bibinfo{booktitle}{\emph{Advances in neural information processing
  systems}}. \bibinfo{pages}{3387--3395}.
\newblock


\bibitem[\protect\citeauthoryear{Ordookhanians, Li, Nakandala, and
  Kumar}{Ordookhanians et~al\mbox{.}}{2019}]%
        {ordookhanians2019demonstration}
\bibfield{author}{\bibinfo{person}{Allen Ordookhanians}, \bibinfo{person}{Xin
  Li}, \bibinfo{person}{Supun Nakandala}, {and} \bibinfo{person}{Arun Kumar}.}
  \bibinfo{year}{2019}\natexlab{}.
\newblock \showarticletitle{Demonstration of Krypton: optimized CNN inference
  for occlusion-based deep CNN explanations}.
\newblock \bibinfo{journal}{\emph{Proceedings of the VLDB Endowment}}
  \bibinfo{volume}{12}, \bibinfo{number}{12} (\bibinfo{year}{2019}),
  \bibinfo{pages}{1894--1897}.
\newblock


\bibitem[\protect\citeauthoryear{Parameswaran}{Parameswaran}{2019}]%
        {parameswaran2019enabling}
\bibfield{author}{\bibinfo{person}{Aditya Parameswaran}.}
  \bibinfo{year}{2019}\natexlab{}.
\newblock \showarticletitle{Enabling data science for the majority}.
\newblock \bibinfo{journal}{\emph{Proceedings of the VLDB Endowment}}
  \bibinfo{volume}{12}, \bibinfo{number}{12} (\bibinfo{year}{2019}),
  \bibinfo{pages}{2309--2322}.
\newblock


\bibitem[\protect\citeauthoryear{Petsiuk, Das, and Saenko}{Petsiuk
  et~al\mbox{.}}{2018}]%
        {petsiuk2018rise}
\bibfield{author}{\bibinfo{person}{Vitali Petsiuk}, \bibinfo{person}{Abir Das},
  {and} \bibinfo{person}{Kate Saenko}.} \bibinfo{year}{2018}\natexlab{}.
\newblock \showarticletitle{Rise: Randomized input sampling for explanation of
  black-box models}.
\newblock \bibinfo{journal}{\emph{arXiv preprint arXiv:1806.07421}}
  (\bibinfo{year}{2018}).
\newblock


\bibitem[\protect\citeauthoryear{Qian, Popa, and Sen}{Qian
  et~al\mbox{.}}{2019}]%
        {qian2019systemer}
\bibfield{author}{\bibinfo{person}{Kun Qian}, \bibinfo{person}{Lucian Popa},
  {and} \bibinfo{person}{Prithviraj Sen}.} \bibinfo{year}{2019}\natexlab{}.
\newblock \showarticletitle{SystemER: a human-in-the-loop system for
  explainable entity resolution}.
\newblock \bibinfo{journal}{\emph{Proceedings of the VLDB Endowment}}
  \bibinfo{volume}{12}, \bibinfo{number}{12} (\bibinfo{year}{2019}),
  \bibinfo{pages}{1794--1797}.
\newblock


\bibitem[\protect\citeauthoryear{Rebuffi, Fong, Ji, and Vedaldi}{Rebuffi
  et~al\mbox{.}}{2020}]%
        {rebuffi2020there}
\bibfield{author}{\bibinfo{person}{Sylvestre-Alvise Rebuffi},
  \bibinfo{person}{Ruth Fong}, \bibinfo{person}{Xu Ji}, {and}
  \bibinfo{person}{Andrea Vedaldi}.} \bibinfo{year}{2020}\natexlab{}.
\newblock \showarticletitle{There and Back Again: Revisiting Backpropagation
  Saliency Methods}. In \bibinfo{booktitle}{\emph{Proceedings of the IEEE/CVF
  Conference on Computer Vision and Pattern Recognition}}.
  \bibinfo{pages}{8839--8848}.
\newblock


\bibitem[\protect\citeauthoryear{Ren, He, Girshick, and Sun}{Ren
  et~al\mbox{.}}{2015}]%
        {ren2015faster}
\bibfield{author}{\bibinfo{person}{Shaoqing Ren}, \bibinfo{person}{Kaiming He},
  \bibinfo{person}{Ross Girshick}, {and} \bibinfo{person}{Jian Sun}.}
  \bibinfo{year}{2015}\natexlab{}.
\newblock \showarticletitle{Faster r-cnn: Towards real-time object detection
  with region proposal networks}. In \bibinfo{booktitle}{\emph{Advances in
  neural information processing systems}}. \bibinfo{pages}{91--99}.
\newblock


\bibitem[\protect\citeauthoryear{Ribeiro, Singh, and Guestrin}{Ribeiro
  et~al\mbox{.}}{2016}]%
        {ribeiro2016should}
\bibfield{author}{\bibinfo{person}{Marco~Tulio Ribeiro},
  \bibinfo{person}{Sameer Singh}, {and} \bibinfo{person}{Carlos Guestrin}.}
  \bibinfo{year}{2016}\natexlab{}.
\newblock \showarticletitle{Why should i trust you?: Explaining the predictions
  of any classifier}. In \bibinfo{booktitle}{\emph{Proceedings of the 22nd ACM
  SIGKDD international conference on knowledge discovery and data mining}}.
  ACM, \bibinfo{pages}{1135--1144}.
\newblock


\bibitem[\protect\citeauthoryear{Russakovsky, Deng, Su, Krause, Satheesh, Ma,
  Huang, Karpathy, Khosla, Bernstein, Berg, and Fei-Fei}{Russakovsky
  et~al\mbox{.}}{2015}]%
        {ILSVRC15}
\bibfield{author}{\bibinfo{person}{Olga Russakovsky}, \bibinfo{person}{Jia
  Deng}, \bibinfo{person}{Hao Su}, \bibinfo{person}{Jonathan Krause},
  \bibinfo{person}{Sanjeev Satheesh}, \bibinfo{person}{Sean Ma},
  \bibinfo{person}{Zhiheng Huang}, \bibinfo{person}{Andrej Karpathy},
  \bibinfo{person}{Aditya Khosla}, \bibinfo{person}{Michael Bernstein},
  \bibinfo{person}{Alexander~C. Berg}, {and} \bibinfo{person}{Li Fei-Fei}.}
  \bibinfo{year}{2015}\natexlab{}.
\newblock \showarticletitle{{ImageNet Large Scale Visual Recognition
  Challenge}}.
\newblock \bibinfo{journal}{\emph{International Journal of Computer Vision
  (IJCV)}} \bibinfo{volume}{115}, \bibinfo{number}{3} (\bibinfo{year}{2015}),
  \bibinfo{pages}{211--252}.
\newblock
\urldef\tempurl%
\url{https://doi.org/10.1007/s11263-015-0816-y}
\showDOI{\tempurl}


\bibitem[\protect\citeauthoryear{Samek, Binder, Montavon, Lapuschkin, and
  M{\"u}ller}{Samek et~al\mbox{.}}{2016}]%
        {samek2016evaluating}
\bibfield{author}{\bibinfo{person}{Wojciech Samek}, \bibinfo{person}{Alexander
  Binder}, \bibinfo{person}{Gr{\'e}goire Montavon}, \bibinfo{person}{Sebastian
  Lapuschkin}, {and} \bibinfo{person}{Klaus-Robert M{\"u}ller}.}
  \bibinfo{year}{2016}\natexlab{}.
\newblock \showarticletitle{Evaluating the visualization of what a deep neural
  network has learned}.
\newblock \bibinfo{journal}{\emph{IEEE transactions on neural networks and
  learning systems}} \bibinfo{volume}{28}, \bibinfo{number}{11}
  (\bibinfo{year}{2016}), \bibinfo{pages}{2660--2673}.
\newblock


\bibitem[\protect\citeauthoryear{Sellam, Lin, Huang, Yang, Vondrick, and
  Wu}{Sellam et~al\mbox{.}}{2019}]%
        {sellam2019deepbase}
\bibfield{author}{\bibinfo{person}{Thibault Sellam}, \bibinfo{person}{Kevin
  Lin}, \bibinfo{person}{Ian Huang}, \bibinfo{person}{Michelle Yang},
  \bibinfo{person}{Carl Vondrick}, {and} \bibinfo{person}{Eugene Wu}.}
  \bibinfo{year}{2019}\natexlab{}.
\newblock \showarticletitle{Deepbase: Deep inspection of neural networks}. In
  \bibinfo{booktitle}{\emph{Proceedings of the 2019 International Conference on
  Management of Data}}. \bibinfo{pages}{1117--1134}.
\newblock


\bibitem[\protect\citeauthoryear{Selvaraju, Cogswell, Das, Vedantam, Parikh,
  and Batra}{Selvaraju et~al\mbox{.}}{2017}]%
        {selvaraju2017grad}
\bibfield{author}{\bibinfo{person}{Ramprasaath~R. Selvaraju},
  \bibinfo{person}{Michael Cogswell}, \bibinfo{person}{Abhishek Das},
  \bibinfo{person}{Ramakrishna Vedantam}, \bibinfo{person}{Devi Parikh}, {and}
  \bibinfo{person}{Dhruv Batra}.} \bibinfo{year}{2017}\natexlab{}.
\newblock \showarticletitle{Grad-{CAM}: Visual explanations from deep networks
  via gradient-based localization}. In \bibinfo{booktitle}{\emph{Proceedings of
  the IEEE International Conference on Computer Vision}}.
  \bibinfo{pages}{618--626}.
\newblock


\bibitem[\protect\citeauthoryear{Simonyan, Vedaldi, and Zisserman}{Simonyan
  et~al\mbox{.}}{2013}]%
        {simonyan2013deep}
\bibfield{author}{\bibinfo{person}{Karen Simonyan}, \bibinfo{person}{Andrea
  Vedaldi}, {and} \bibinfo{person}{Andrew Zisserman}.}
  \bibinfo{year}{2013}\natexlab{}.
\newblock \showarticletitle{Deep inside convolutional networks: Visualising
  image classification models and saliency maps}.
\newblock \bibinfo{journal}{\emph{arXiv preprint arXiv:1312.6034}}
  (\bibinfo{year}{2013}).
\newblock


\bibitem[\protect\citeauthoryear{Simonyan and Zisserman}{Simonyan and
  Zisserman}{2014}]%
        {simonyan2014very}
\bibfield{author}{\bibinfo{person}{Karen Simonyan} {and}
  \bibinfo{person}{Andrew Zisserman}.} \bibinfo{year}{2014}\natexlab{}.
\newblock \showarticletitle{Very deep convolutional networks for large-scale
  image recognition}.
\newblock \bibinfo{journal}{\emph{arXiv preprint arXiv:1409.1556}}
  (\bibinfo{year}{2014}).
\newblock


\bibitem[\protect\citeauthoryear{Springenberg, Dosovitskiy, Brox, and
  Riedmiller}{Springenberg et~al\mbox{.}}{2014}]%
        {springenberg2014striving}
\bibfield{author}{\bibinfo{person}{Jost~Tobias Springenberg},
  \bibinfo{person}{Alexey Dosovitskiy}, \bibinfo{person}{Thomas Brox}, {and}
  \bibinfo{person}{Martin Riedmiller}.} \bibinfo{year}{2014}\natexlab{}.
\newblock \showarticletitle{Striving for simplicity: The all convolutional
  net}.
\newblock \bibinfo{journal}{\emph{arXiv preprint arXiv:1412.6806}}
  (\bibinfo{year}{2014}).
\newblock


\bibitem[\protect\citeauthoryear{Sundararajan, Taly, and Yan}{Sundararajan
  et~al\mbox{.}}{[n. d.]}]%
        {sundararajan1703axiomatic}
\bibfield{author}{\bibinfo{person}{M Sundararajan}, \bibinfo{person}{A Taly},
  {and} \bibinfo{person}{Q Yan}.} \bibinfo{year}{[n. d.]}\natexlab{}.
\newblock \showarticletitle{Axiomatic attribution for deep networks. arXiv
  2017}.
\newblock \bibinfo{journal}{\emph{arXiv preprint arXiv:1703.01365}}
  (\bibinfo{year}{[n. d.]}).
\newblock


\bibitem[\protect\citeauthoryear{Vartak, F.~da Trindade, Madden, and
  Zaharia}{Vartak et~al\mbox{.}}{2018}]%
        {vartak2018mistique}
\bibfield{author}{\bibinfo{person}{Manasi Vartak}, \bibinfo{person}{Joana~M
  F.~da Trindade}, \bibinfo{person}{Samuel Madden}, {and}
  \bibinfo{person}{Matei Zaharia}.} \bibinfo{year}{2018}\natexlab{}.
\newblock \showarticletitle{Mistique: A system to store and query model
  intermediates for model diagnosis}. In \bibinfo{booktitle}{\emph{Proceedings
  of the 2018 International Conference on Management of Data}}.
  \bibinfo{pages}{1285--1300}.
\newblock


\bibitem[\protect\citeauthoryear{Voulodimos, Doulamis, Doulamis, and
  Protopapadakis}{Voulodimos et~al\mbox{.}}{2018}]%
        {voulodimos2018deep}
\bibfield{author}{\bibinfo{person}{Athanasios Voulodimos},
  \bibinfo{person}{Nikolaos Doulamis}, \bibinfo{person}{Anastasios Doulamis},
  {and} \bibinfo{person}{Eftychios Protopapadakis}.}
  \bibinfo{year}{2018}\natexlab{}.
\newblock \showarticletitle{Deep learning for computer vision: A brief review}.
\newblock \bibinfo{journal}{\emph{Computational intelligence and neuroscience}}
   \bibinfo{volume}{2018} (\bibinfo{year}{2018}).
\newblock


\bibitem[\protect\citeauthoryear{Yang, Du, and Hu}{Yang et~al\mbox{.}}{2019}]%
        {yang2019evaluating}
\bibfield{author}{\bibinfo{person}{Fan Yang}, \bibinfo{person}{Mengnan Du},
  {and} \bibinfo{person}{Xia Hu}.} \bibinfo{year}{2019}\natexlab{}.
\newblock \showarticletitle{Evaluating explanation without ground truth in
  interpretable machine learning}.
\newblock \bibinfo{journal}{\emph{arXiv preprint arXiv:1907.06831}}
  (\bibinfo{year}{2019}).
\newblock


\bibitem[\protect\citeauthoryear{Yang and Kim}{Yang and Kim}{2019}]%
        {yang2019bim}
\bibfield{author}{\bibinfo{person}{Mengjiao Yang} {and} \bibinfo{person}{Been
  Kim}.} \bibinfo{year}{2019}\natexlab{}.
\newblock \showarticletitle{BIM: Towards quantitative evaluation of
  interpretability methods with ground truth}.
\newblock \bibinfo{journal}{\emph{arXiv preprint arXiv:1907.09701}}
  (\bibinfo{year}{2019}).
\newblock


\bibitem[\protect\citeauthoryear{Yeh, Hsieh, Suggala, Inouye, and
  Ravikumar}{Yeh et~al\mbox{.}}{2019}]%
        {yeh2019fidelity}
\bibfield{author}{\bibinfo{person}{Chih-Kuan Yeh}, \bibinfo{person}{Cheng-Yu
  Hsieh}, \bibinfo{person}{Arun Suggala}, \bibinfo{person}{David~I Inouye},
  {and} \bibinfo{person}{Pradeep~K Ravikumar}.}
  \bibinfo{year}{2019}\natexlab{}.
\newblock \showarticletitle{On the (in) fidelity and sensitivity of
  explanations}. In \bibinfo{booktitle}{\emph{Advances in Neural Information
  Processing Systems}}. \bibinfo{pages}{10967--10978}.
\newblock


\bibitem[\protect\citeauthoryear{Zeiler and Fergus}{Zeiler and Fergus}{2014}]%
        {zeiler2014visualizing}
\bibfield{author}{\bibinfo{person}{Matthew~D Zeiler} {and} \bibinfo{person}{Rob
  Fergus}.} \bibinfo{year}{2014}\natexlab{}.
\newblock \showarticletitle{Visualizing and understanding convolutional
  networks}. In \bibinfo{booktitle}{\emph{European conference on computer
  vision}}. Springer, \bibinfo{pages}{818--833}.
\newblock


\bibitem[\protect\citeauthoryear{Zhang, Bargal, Lin, Brandt, Shen, and
  Sclaroff}{Zhang et~al\mbox{.}}{2018}]%
        {zhang2018top}
\bibfield{author}{\bibinfo{person}{Jianming Zhang}, \bibinfo{person}{Sarah~Adel
  Bargal}, \bibinfo{person}{Zhe Lin}, \bibinfo{person}{Jonathan Brandt},
  \bibinfo{person}{Xiaohui Shen}, {and} \bibinfo{person}{Stan Sclaroff}.}
  \bibinfo{year}{2018}\natexlab{}.
\newblock \showarticletitle{Top-down neural attention by excitation backprop}.
\newblock \bibinfo{journal}{\emph{International Journal of Computer Vision}}
  \bibinfo{volume}{126}, \bibinfo{number}{10} (\bibinfo{year}{2018}),
  \bibinfo{pages}{1084--1102}.
\newblock


\bibitem[\protect\citeauthoryear{Zhou, Khosla, Lapedriza, Oliva, and
  Torralba}{Zhou et~al\mbox{.}}{2016}]%
        {zhou2016learning}
\bibfield{author}{\bibinfo{person}{Bolei Zhou}, \bibinfo{person}{Aditya
  Khosla}, \bibinfo{person}{Agata Lapedriza}, \bibinfo{person}{Aude Oliva},
  {and} \bibinfo{person}{Antonio Torralba}.} \bibinfo{year}{2016}\natexlab{}.
\newblock \showarticletitle{Learning deep features for discriminative
  localization}. In \bibinfo{booktitle}{\emph{Proceedings of the IEEE
  conference on computer vision and pattern recognition}}.
  \bibinfo{pages}{2921--2929}.
\newblock


\bibitem[\protect\citeauthoryear{Zhou, Lapedriza, Khosla, Oliva, and
  Torralba}{Zhou et~al\mbox{.}}{2017}]%
        {zhou2017places}
\bibfield{author}{\bibinfo{person}{Bolei Zhou}, \bibinfo{person}{Agata
  Lapedriza}, \bibinfo{person}{Aditya Khosla}, \bibinfo{person}{Aude Oliva},
  {and} \bibinfo{person}{Antonio Torralba}.} \bibinfo{year}{2017}\natexlab{}.
\newblock \showarticletitle{Places: A 10 million Image Database for Scene
  Recognition}.
\newblock \bibinfo{journal}{\emph{IEEE Transactions on Pattern Analysis and
  Machine Intelligence}} (\bibinfo{year}{2017}).
\newblock


\end{thebibliography}

%

\end{document}